\pdfoutput=1

\documentclass[11pt]{article}

\usepackage[final]{acl}

\usepackage{xcolor, colortbl}
\usepackage{array}
\usepackage{booktabs}
\usepackage{enumitem}
\usepackage{graphicx}
\usepackage{hyperref}
\usepackage{latexsym}
\usepackage{lipsum}
\usepackage{multicol}
\usepackage{multirow}
\usepackage{subcaption}
\usepackage{tabularx}
\usepackage{tabularray}
\usepackage{times}
\usepackage{todonotes}
\usepackage{xspace}

\usepackage[T1]{fontenc}

\usepackage[utf8]{inputenc}

\usepackage{microtype}

\usepackage{inconsolata}

\newcommand{\benchmark}{\textsc{MM-Soc}\xspace}

\title{\benchmark: Benchmarking Multimodal Large Language Models in Social Media Platforms}

\author{
    Yiqiao Jin\textsuperscript{\rm 1}, 
    Minje Choi\textsuperscript{\rm 1}, 
    Gaurav Verma\textsuperscript{\rm 1}, 
    Jindong Wang\textsuperscript{\rm 2}, 
    Srijan Kumar\textsuperscript{\rm 1} 
    \\
  \textsuperscript{\rm 1}Georgia Institute of Technology \\
  \textsuperscript{\rm 2}Microsoft Research Asia \\
  \texttt{\{yjin328,mchoi96,gverma,srijan\}@gatech.edu} \\
  \texttt{jindong.wang@microsoft.com}}

\begin{document}
\maketitle
\begin{abstract}
Social media platforms are hubs for multimodal information exchange, encompassing text, images, and videos, making it challenging for machines to comprehend the information or emotions associated with interactions in online spaces. Multimodal Large Language Models (MLLMs) have emerged as a promising solution to these challenges, yet they struggle to accurately interpret human emotions and complex content such as misinformation. This paper introduces \benchmark, a comprehensive benchmark designed to evaluate MLLMs' understanding of multimodal social media content. \benchmark compiles prominent multimodal datasets and incorporates a novel large-scale YouTube tagging dataset, targeting a range of tasks from misinformation detection, hate speech detection, and social context generation. 
Through our exhaustive evaluation on ten size-variants of four open-source MLLMs, we have identified significant performance disparities, highlighting the need for advancements in models' social understanding capabilities. 
Our analysis reveals that, in a zero-shot setting, various types of MLLMs generally exhibit difficulties in handling social media tasks. However, MLLMs demonstrate performance improvements post fine-tuning, suggesting potential pathways for improvement. Our code and data are available at \url{https://github.com/claws-lab/MMSoc.git}. 
\end{abstract}

\section{Introduction}
\label{sec:intro}

\begin{figure}[t]
    \centering
    \includegraphics[width=0.98\linewidth]{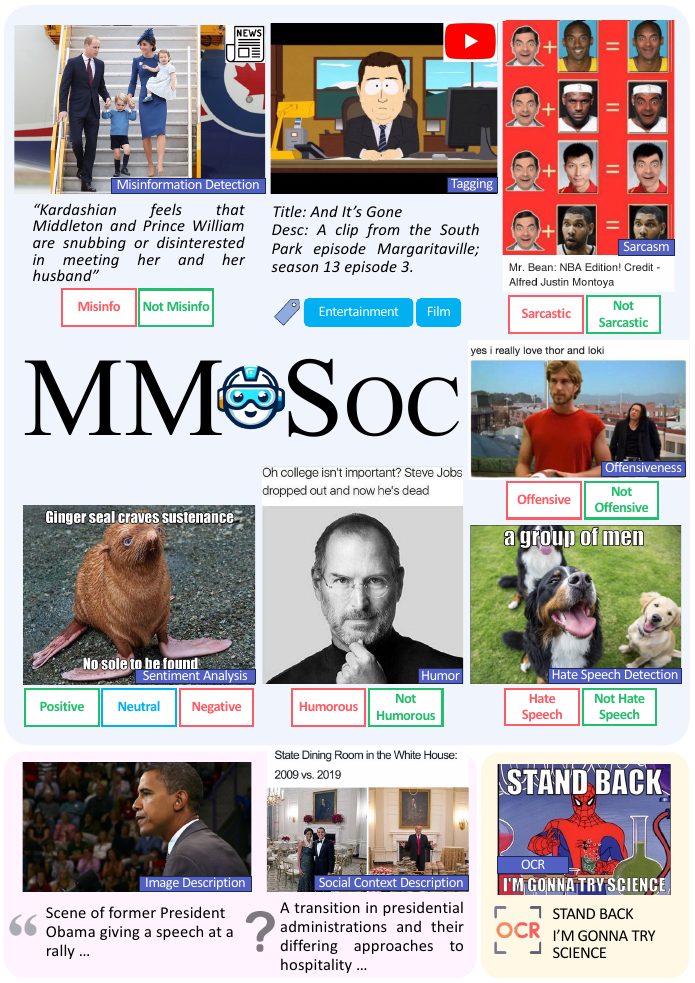}
    \vspace{-1mm}
    \caption{The 
    \benchmark benchmark includes 10 multimodal tasks, including 7 image-text classification tasks (misinformation detection, tagging, sarcasm, offensiveness, sentiment analysis, hate speech detection, and humor), 2 generative task (image description and social context description) and a text extraction task (OCR).
    }
    \label{fig:pipeline}
    \vspace{-4mm}
\end{figure}
Social media platforms have become the epicenter of multimodal information exchange, blending various formats of content such as text, images, and videos. These platforms serve not only as channels for sharing news and personal experiences but also for spreading rumors and shaping public opinion~\cite{ferrara2020dynamics,vosoughi2018spread}. The inherent multimodality of social media content requires users to not only interpret individual modalities such as text or images but also to understand the interplay between them, pushing the boundaries of how machines comprehend human communication in online spaces.

Multimodal Large Language Models (MLLMs) have recently emerged as powerful tools for bridging the understanding of natural language and visual cues, showcasing their potential in a range of tasks ranging from image captioning to complex question answering~\cite{ramos2023captioning,liu2023visual,liu2023improved}. Despite these advancements, 
%
the complexity of tasks such as understanding human emotions, memes, and verifying misinformation presents significant evaluation challenges to MLLMs~\cite{chen2023can, chen2023combating}. These tasks require not only combining signals extracted from both textual and visual domains, \textit{but} also considering various social contexts upon making a decision regarding contextual appropriateness or correctness, which often require knowledge of cultural contexts and subjective interpretations~\cite{ruch2010humor,jacobi2014perceptions}. For instance, the task of explaining visual memes requires not only proficiency in image recognition and language generation, but also capability of understanding the underlying situation of the image on why it should be considered humorous. Given that large language models struggle at solving tasks requiring social knowledge~\cite{choi2023llms}, we anticipate multimodal social tasks to prove an even harder challenge.

The complexity of multimodal tasks from social media demands a benchmark that can evaluate MLLMs on their understanding of the different data domains as well as the social context. Such a benchmark would not only highlight the current limitations of MLLMs, but also lead to future innovations aimed at bridging the gap between human and machine understanding of multimodal content.

\noindent \textbf{This Work.} This paper introduces \benchmark, a novel multimodal benchmark to rigorously assess the capabilities of MLLMs across diverse tasks typical of social media environments.
Along with existing prominent multimodal datasets, we add a large-scale, newly collected YouTube tagging dataset, resulting in ten tasks across five datasets. 
Our analysis primarily targets open-source MLLMs,  
recognizing their advantages in terms of rapid deployment, reduced operational costs, and superior capacity for maintaining data integrity compared to centralized proprietary models. 
Through \benchmark, we conduct a thorough and systematic examination of MLLMs, exploring and validating new methodologies to augment MLLM efficacy in handling multimodal tasks. Finally, we provide a detailed discussion on the performances, shedding light on the implications of our findings for future MLLM development and deployment.

\noindent \textbf{Contributions.} Our contributions are summarized as follows. First, we introduce \benchmark, a novel benchmark to holistically evaluate MLLMs’ capability in tackling multimodal tasks derived from online social networks.
Second, we perform a comprehensive evaluation and benchmark 10 representative open-source MLLMs on \benchmark, comparing their performances with fine-tuned LLM baselines.
Third, we conduct two case studies on \benchmark for testing the effectiveness of two methods: self-improvement and explanation-augmented finetuning.
We find that, while zero-shot MLLMs often fall short in achieving comparable performances compared to fine-tuned models, their performances can be improved via specific fine-tuning strategies. 


%

\section{The \benchmark Benchmark}
\label{sec:dataset}
\vspace{-2mm}


\noindent \textbf{Overview.} 
The deployment of Multimodal Large Language Models (MLLMs) as general-purpose assistants across social networks marks a significant shift from traditional, specialized models designed for singular tasks. 
This transition necessitates a comprehensive skill set enabling these models to navigate the multifaceted challenges presented by user-generated content. 

Motivated by this, we design \benchmark, which spans both natural language understanding and generation tasks. These tasks are designed to test the models' abilities to interact with user-generated content encountered online. The selection includes binary classification, multi-class classification, text extraction, and text generation tasks, aiming to cover a wide range of interactions MLLMs might encounter with online content. The detailed task selection process is in Appendix~\ref{sec:task_selection}. 
To ensure a comprehensive evaluation, we employ a variety of 10 tasks that mirror the complexity of real-world scenarios, ranging from understanding online video content to identifying misinformation and detecting hate speech in memes. 
The statistics of the dataset are in Table~\ref{tab:dataset}. 

\noindent \textbf{Tagging.} In digital content management, the ability to accurately predict appropriate tags for online content is particularly significant given their diverse and multimodal nature, which includes textual narratives, visual features, and cultural contexts. 
Effective tagging enhances content discoverability, facilitates content moderation, and significantly improves the user experience. 
To this end, we introduce \emph{YouTube2M}, a novel dataset comprising 2 million YouTube videos shared on Reddit, specifically curated to assess models' proficiency in predicting tags from a predefined set in Table~\ref{tab:YouTubeTags} based on video titles, descriptions, and visual content. We provide a comprehensive analysis of \emph{YouTube2M} in Appendix~\ref{app:YouTube2M}. 

\emph{YouTube2M} distinguishes itself with two features:  
\emph{1) Relevance to Online User Groups.} The \emph{YouTube2M} dataset features videos shared on Reddit. The selection of YouTube as the primary source is based on its expansive user base, with more than 2.5 billion monthly users~\cite{YouTubeMonthlyActiveUsers} and the rich variety of its multimodal content. Reddit is among the top most popular social media and is characterized by its unique community structures called ``subreddits''. Unlike general video collections on YouTube, \emph{YouTube2M} reflects the choices of individuals within specific subreddit communities, aligning with their interests, humor, or preferences. This targeted selection process ensures our dataset is particularly relevant to distinct user groups. 
\emph{2) Viral Potential.} Reddit is renowned as a catalyst for virality. Videos shared on Reddit can rapidly gain significant attention and engage communities deeply through more discussions, comments, and votes within their respective subreddits. 
Notably, the presence of toxic, biased, or unverified content in online videos raises concerns over the propagation of misinformation, fostering distrust and hate speech online. Consequently, the accurate categorization and tagging of these videos become critical for content moderation. 


\noindent \underline{\emph{Dataset Construction.}} 
We retrieved the URLs of all YouTube videos shared on Reddit over 12 years spanning from 2011 to 2022. Subsequently, we used YouTube Data API~\footnote{\url{https://developers.google.com/youtube/v3}} to collect comprehensive metadata of the YouTube videos, including their titles, descriptions, channels, publish timestamps, restrictions, default languages, topic categories, and embeddability status. 
Additionally, we compiled extensive statistics for each video, covering aspects such as duration, and the number of comments, likes, and views they garnered. 
To ensure the quality and relevance of the dataset, we filtered the dataset and retained only videos with valid tags and thumbnail images, resulting in a dataset with 1,963,697 videos.




\noindent \textbf{Misinformation Detection.} Misinformation detection represents a critical challenge as the proliferation of multimodal misinformation across online platforms can undermine trust in digital ecosystems and lead to real-world harm~\cite{yang2022reinforcement, yang2023multi, ma2022curriculum, jin2022towards, he2023survey}. Here, we formulate misinformation detection as a binary classification problem and utilize the \underline{PolitiFact} and \underline{GossipCop} datasets~\cite{shu2020fakenewsnet}. The task aims at evaluating a model's ability to accurately differentiate between true news and misinformation, leveraging both the textual content and the associated images of news articles. 

\noindent \textbf{Hate Speech Detection.} The prevalence of hate speech in online platforms has several detrimental effects, both on the individual user-level and on the platform as a whole~\cite{mondal2017measurement, he2021racism}.
To support research targeted at curbing the spread of harmful content and abusive language, we incorporate the \underline{Hateful Memes}~\cite{kiela2020hateful} dataset. 
This dataset evaluates the ability to recognize messages that attack or demean a group based on attributes such as race, religion, ethnic origin, sexual orientation, disability, or gender. Such ability is essential for creating inclusive online environments, protecting users from harm, and complying with legal standards.



\noindent \textbf{Emotion Analysis.} The interactions among users in online social media platforms often contain rich and diverse exchanges of emotions. These emotions include not only sentiment but also humor, sarcasm, and offensiveness. Coupled with multimodal means of expressions such as memes, it can be challenging for MLLMs to accurately capture the true emotion conveyed through the message.
Therefore, we include the \underline{Memotion}~\cite{sharma2020semeval} dataset which focuses on sentiment and emotion analysis within online memes, presenting a multifaceted challenge that spans sentiment analysis and the detection of humor, sarcasm, and offensive contents.

\noindent \textbf{OCR.} Optical character recognition (OCR) refers to the task of extracting text within images into machine-encoded text. 
A model's OCR proficiency is directly related to its ability to access and interpret online information such as infographics, memes, and screenshots of textual conversations, which are prevalent forms of communication and information dissemination online~\cite{zannettou2018memes}. We use the Hateful Memes and Memotion datasets to evaluate OCR capabilities.

\noindent \textbf{Image \& Social Context Description.} Image description assesses a model's ability to generate accurate, contextually relevant, and coherent natural language descriptions of images. The capability to accurately describe an image in natural language aids in the understanding of the visual content, which both provides an intermediary step in reasoning about the multimodal inputs and also aids human users in understanding their decisions in an interpretable way.
Previous studies have demonstrated that commercial models such as GPT-4/3.5 possess extensive domain knowledge in various fields, including social sciences, and have shown promising results in data annotation, surpassing the performance of human annotators~\cite{savelka2023can, gilardi2023chatgpt, zhu2023minigpt}. Thus, for each example in the dataset, we employed GPT-4V as a strong teacher to generate descriptions of images and their associated social contexts. For each example within the dataset, we instructed the model to provide a comprehensive description of the image, encompassing its foreground, background, major subjects, colors, and textures, as well as the social context for each example, such as cultural backgrounds, possible interpretations within various societal groups, and the potential target demographics. These examples served as references for evaluating MLLMs' capabilities to understand both the image contents and social knowledge. 






\subsection{Model Selection}
\label{sec:model_selection}

\begin{table*}[!ht]
\vspace{-3mm}
    \centering
    \small
    \begin{tabular}{l|cccccccccc|c}
    \toprule
         Model & Misinfo & Hate & Humor & Sarc. & Off. & Sent. & Tag & OCR & ID & SCD & Avg. \\ 
        \midrule
        llava-v1.5-7b & 0.494 & 0.490 & 0.450 & 0.452 & \textbf{0.484} & 0.250 & 0.068 & 0.514 & \textbf{0.260} & \textbf{0.218} & \underline{0.368} \\ 
        llava-v1.5-13b & \textbf{0.642} & \underline{0.578} & \underline{0.534} & 0.436 & 0.451 & 0.291 & 0.071 & 0.542 & \underline{0.259} & \underline{0.216} & \textbf{0.402} \\ 
        instructblip-vicuna-7b & 0.311 & 0.442 & 0.246 & 0.481 & \underline{0.477} & 0.251 & / & 0.611 & 0.048 & 0.033 & 0.322 \\ 
        instructblip-vicuna-13b & 0.435 & 0.528 & 0.435 & 0.437 & 0.417 & 0.262 & 0.050 & 0.701 & 0.097 & 0.020 & 0.338 \\ 
        instructblip-flan-t5-xl & 0.455 & 0.470 & 0.282 & 0.274 & 0.464 & 0.185 & 0.057 & 0.652 & 0.041 & 0.046 & 0.293 \\ 
        instructblip-flan-t5-xxl & 0.463 & 0.570 & 0.406 & 0.447 & 0.282 & \textbf{0.335} & 0.128 & 0.627 & 0.043 & 0.023 & 0.332 \\ 
        blip2-opt-2.7b & 0.261 & 0.369 & 0.309 & 0.389 & 0.411 & 0.291 & 0.022 & \textbf{0.723} & 0.141 & 0.140 & 0.306 \\ 
        blip2-flan-t5-xl & 0.467 & 0.400 & 0.183 & \underline{0.497} & 0.282 & 0.245 & \underline{0.157} & \underline{0.718} & 0.147 & 0.137 & 0.323 \\ 
        blip2-flan-t5-xxl & 0.373 & \textbf{0.587} & 0.200 & \textbf{0.512} & 0.282 & \underline{0.295} & \textbf{0.188} & 0.676 & 0.133 & 0.113 & 0.336 \\ 
        llama-adapter-v2 & \underline{0.553} & 0.524 & \textbf{0.556} & 0.453 & 0.471 & 0.268 & 0.021 & 0.111 & 0.098 & 0.139 & 0.319 \\ 
        \midrule
        random & 0.459 & 0.500 & 0.467 & 0.460 & 0.493 & 0.286 & / & / & / & / & / \\ 
        \bottomrule
    \end{tabular}
    \vspace{-2mm}
    \caption{Performance comparison across all models on the tasks. Best and 2nd best performances among the MLLMs are highlighted in \textbf{bold} and \underline{underline}, respectively. ``ID'' and ``SCD'' stand for the image description task and the social context description task, respectively. 
    Note that instructblip-vicuna-7b fails to generate valid answers on the tagging task. 
    A full comparison of all models on all metrics can be found in Appendix~\ref{app:fullEvalResults}. 
    }
    \label{tab:overallF1}
\end{table*}

We consider 10 prominent open-source models spanning four different distinct architectures: LLaVA-v1.5~\cite{liu2023improved}, BLIP2~\cite{li2023blip}, InstructBLIP~\cite{dai2023instructblip}, and LLaMA-Adapter-v2~\cite{zhang2023llama}. 
Details on model parameter volumes are in Table~\ref{tab:models}.
The models are selected to cover diverse model sizes. We apply our prompts (Table~\ref{tab:prompt}) to test the performances of MLLMs in a zero-shot setting. 
For tasks in which ground-truth texts are available as inputs, we compare MLLMs' performances with five unimodal discriminative models in a full fine-tuning setting, including BERT~\cite{kenton2019bert}, RoBERTa-Base/Large~\cite{liu2019roberta}, DeBERTa~\cite{he2020deberta}, and MiniLM~\cite{wang2020minilm}. 
These text-only models have shown competitive performances in text classification. Implementation details can be found in Appendix~\ref{app:implementation_details}.

\section{Benchmark Results}
\label{sec:bench}
\vspace{-2mm}
Table~\ref{tab:overallF1} shows the overall performances across 10 tasks. Here, we use a unified score for each task to facilitate a high-level performance comparison across diverse tasks.
For text classification and extraction tasks, we use the macro-F1 score as the aggregated measure. 
For text generation tasks including image description (ID) and social context description (SCD), we use ROUGE-L~\cite{lin2004rouge}. The results for misinformation detection are averaged across PolitiFact and GossipCop, and the results for OCR are averaged across Memotion and Hateful Memes. The complete evaluation results can be found in Appendix~\ref{app:fullEvalResults}.

\noindent \textbf{Zero-shot MLLMs are on par with random guesses.}
Despite their large model sizes and extensive training corpus, all MLLMs demonstrate underwhelming performances in zero-shot settings, often paralleling and sometimes falling short of the random baseline. 
This trend is especially evident on the offensiveness detection task, where none of the 10 models surpass the random baseline, with an average macro F1 score of 0.402 compared to the baseline of 0.493. A similar pattern emerges in humor detection, with eight models underperforming the baseline. 
The tasks in our benchmark which simulate real-life interactions in social media are indeed challenging for most MLLMs.

\begin{table*}[!ht]
    \centering
    \small
    \begin{tabular}{cl|cccc|cccc}
        \toprule
        ~ & ~ & \multicolumn{4}{c|}{PolitiFact} & \multicolumn{4}{c}{GossipCop} \\
        Setting & Model & $\mathrm{F1}_{\mathrm{macro}}$ & Acc & AUC & SR\% & $\mathrm{F1}_{\mathrm{macro}}$  & Acc & AUC & SR\% \\ 
        \midrule
        \multirow{10}{*}{zero-shot} & llava-v1.5-7b & 0.488 & \underline{0.740} & 0.534 & 100.0 & 0.499 & \underline{0.812} & 0.524 & 100.0 \\ 
        ~ & llava-v1.5-13b & \textbf{0.749} & \textbf{0.827} & \textbf{0.721} & 100.0 & \underline{0.534} & 0.773 & \underline{0.535} & 100.0 \\ 
        ~ & instructblip-vicuna-7b & 0.376 & 0.388 & 0.511 & 76.9 & 0.246 & 0.251 & 0.466 & 70.5 \\ 
        ~ & instructblip-vicuna-13b & 0.434 & 0.485 & 0.441 & 94.2 & 0.435 & 0.503 & 0.468 & 90.0 \\ 
        ~ & instructblip-flan-t5-xl & 0.418 & 0.718 & 0.500 & 99.0 & 0.492 & 0.811 & 0.521 & 98.1 \\
        ~ & instructblip-flan-t5-xxl & 0.519 & 0.543 & 0.537 & 100.0 & 0.406 & 0.429 & 0.497 & 100.0 \\ 
        ~ & blip2-opt-2.7b & 0.213 & 0.227 & 0.429 & 21.2 & 0.309 & 0.309 & 0.437 & 11.2 \\ 
        ~ & blip2-flan-t5-xl & 0.419 & 0.721 & 0.500 & 100.0 & 0.514 & \textbf{0.819} & 0.534 & 100.0 \\
        ~ & blip2-flan-t5-xxl & 0.545 & 0.548 & \underline{0.634} & 100.0 & 0.200 & 0.215 & 0.481 & 100.0 \\ 
        ~ & llama-adapter-v2 & \underline{0.550} & 0.553 & 0.613 & 87.5 & \textbf{0.556} & 0.673 & \textbf{0.581} & 83.6 \\ 
        \midrule
        \multirow{5}{*}{finetuned} & bert-base-uncased & 0.850 & 0.875 & 0.850 & 100.0 & 0.769 & 0.869 & 0.797 & 100.0 \\ 
        ~ & roberta-base & \underline{0.894} & \underline{0.923} & \underline{0.894} & 100.0 & 0.812 & \underline{0.879} & \textbf{0.824} & 100.0 \\ 
        ~ & roberta-large & 0.846 & 0.885 & 0.825 & 100.0 & \textbf{0.820} & 0.858 & \underline{0.820} & 100.0 \\
        ~ & MiniLM-v2 & 0.793 & 0.827 & 0.806 & 100.0 & 0.777 & 0.858 & 0.785 & 100.0 \\ 
        ~ & deberta-v3-large & \textbf{0.952} & \textbf{0.962} & \textbf{0.952} & 100.0 & \underline{0.817} & \textbf{0.895} & 0.792 & 100.0 \\ 
        \midrule
        random & / & 0.471 & 0.500 & 0.494 & / & 0.448 & 0.500 & 0.500 & / \\
    \bottomrule
    \end{tabular}
    \vspace{-2mm}
    \caption{Results of fine-tuning and zero-shot misinformation detection on PolitiFact and GossipCop~\cite{shu2020fakenewsnet}. The best and 2nd best performances of each category is highlighted in \textbf{bold} and \underline. We report the Macro F1-score (F1), Accuracy (Acc), Area Under the Curve (AUC), and Success Rate (SR). As the number of parameters in the model increases, the model is better at following instructions as seen from their increasing success rate. 
    }
    \label{tab:misinfo}
    \vspace{-3mm}
\end{table*}

\noindent \textbf{Zero-shot MLLMs underperform fully finetuned models in most settings.}
We next focus on the misinformation detection task, which takes a binary classification form and can thus be evaluated using encoder-only LLMs such as BERT. Table~\ref{tab:misinfo} reveals a consistent underperformance of MLLMs compared to fully fine-tuned LLMs which \textit{only} use textual information. 
To our surprise, DeBERTa emerges as the top-performing model with only 98 million parameters, whereas zero-shot MLLMs achieve significantly inferior performances. 

The low performances of zero-shot MLLMs can be attributed primarily to two reasons: 
1) \textbf{The divergence in training objectives.} 
Unlike discriminative models, which are explicitly fine-tuned to predict correct labels, MLLMs are oriented towards maximizing cross-modal alignment and instruction-following abilities. Their training regimes are designed to enhance text generation capabilities based on input images. 
Such an alignment does not cater to misinformation detection, which demands not only multimodal reasoning but also the ability to evaluate the reliability of sources and incorporate extensive external knowledge. 
2) \textbf{Disparity in the training corpus content.} 
MLLMs are predominantly trained for tasks such as object detection, image captioning and visual question answering (VQA)~\cite{dai2023instructblip, liu2023visual}, which  
rarely encompass tasks in social knowledge reasoning. The lack of tasks requiring subjective reasoning may inherently limit the MLLMs' performance regarding these tasks, and is further supported by the fact that performing task-specific fine-tuning on even much smaller models that use only limited information significantly outperforms MLLMs.

\begin{figure}[t]
    \centering
    \includegraphics[width=0.75\linewidth]{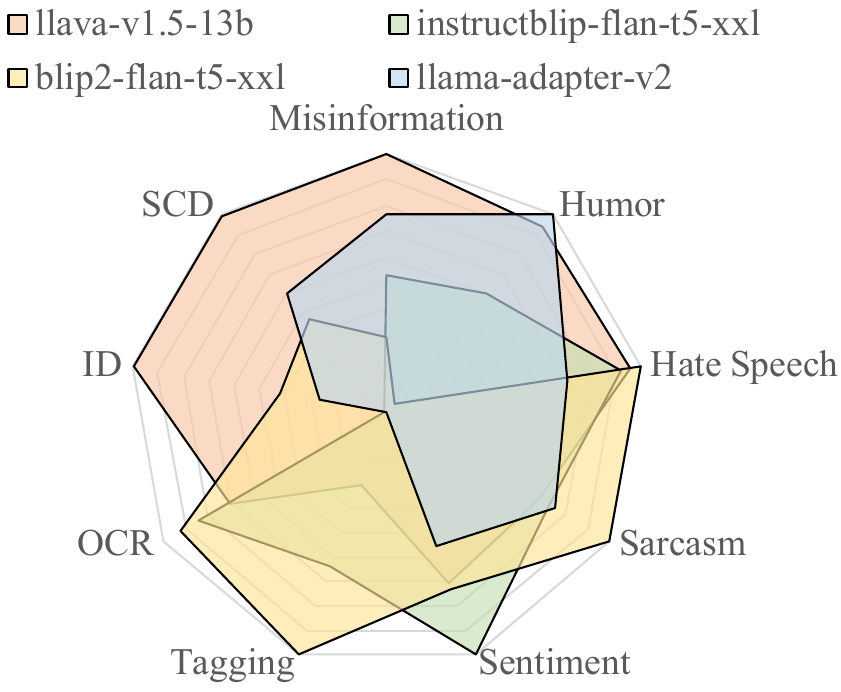}
    \vspace{-3mm}
    \caption{Performances of the 4 representative models on the \benchmark benchmark.
    }
    \label{fig:radar_plot}
    \vspace{-4mm}
\end{figure}

\noindent \textbf{LLaVA achieves highest performance among all MLLMs in most tasks.}
Among the tested MLLMs, LLaVA-v1.5-13b/7b achieve the best and second best overall performances with average scores of 0.402 / 0.368, a 18.9\% / 8.9\% improvement over InstructBLIP Vicuna 13B. The performance gap is most significant on the text generation tasks, including ID and SCD as shown in Table~\ref{tab:overallF1}, where LLaVA-v1.5-13B reaches a performance improvement of 76.9\% and 55.7\% compared with the other models. This advantage could result from both having a wider range of training data and pretraining objectives — multiturn conversation, detailed description, and complex reasoning. 
For example, the complex reasoning objective typically requires a step-by-step reasoning process by following rigorous logic. Figure~\ref{fig:radar_plot} shows the performances of the strongest models under each model architecture. The scores are normalized in the 0-1 range. Interestingly, we found that no single model achieves the best performance across all tasks. LLaVA-v1.5-13B performs the best on text generation such as ID or SCD as well as tasks that require social reasoning like misinformation detection, but its ability in tagging is relatively poor. BLIP2 is best on OCR and discriminative tasks like sarcasm and hate speech detection, whereas its generative abilities are relatively poor. 

\begin{table*}[!ht]
    \centering
    \small
    \begin{tabular}{l|ccccc|ccccc}
    \toprule
         & \multicolumn{5}{c|}{Image Description} & \multicolumn{5}{c|}{Social Context Description} \\
        Model & M & R-1 & R-2 & R-L & Len & M & R-1 & R-2 & R-L & Len \\ 
        \midrule
        instructblip-vicuna-7b & 0.016 & 0.053 & 0.008 & 0.048 & 3.0 & 0.014 & 0.034 & 0.007 & 0.033 & 1.7 \\ 
        instructblip-vicuna-13b & 0.040 & 0.113 & 0.020 & 0.097 & 6.6 & 0.010 & 0.021 & 0.002 & 0.020 & 1.9 \\ 
        instructblip-flan-t5-xl & 0.014 & 0.044 & 0.005 & 0.041 & 2.7 & 0.022 & 0.050 & 0.006 & 0.046 & 3.0 \\ 
        instructblip-flan-t5-xxl & 0.014 & 0.048 & 0.005 & 0.043 & 2.5 & 0.009 & 0.023 & 0.003 & 0.023 & 1.6 \\ 
        blip2-opt-2.7b & 0.076 & 0.158 & 0.025 & 0.141 & 21.2 & 0.081 & 0.163 & 0.021 & 0.140 & 16.3 \\ 
        blip2-flan-t5-xl & 0.065 & 0.172 & 0.026 & 0.147 & 9.8 & 0.069 & 0.156 & 0.024 & 0.137 & 9.5 \\ 
        blip2-flan-t5-xxl & 0.058 & 0.151 & 0.025 & 0.133 & 9.7 & 0.066 & 0.132 & 0.014 & 0.113 & 10.4 \\ 
        llama-adapter-v2 & 0.041 & 0.110 & 0.019 & 0.098 & 9.1 & 0.113 & 0.152 & 0.020 & 0.139 & 128.5 \\ 
        \midrule
        llava-v1.5-7b & 0.223 & 0.288 & 0.074 & 0.260 & 78.2 & 0.229 & 0.247 & 0.057 & 0.218 & 110.1 \\ 
        \quad + FT & 0.217 & 0.285 & 0.074 & 0.253 & 85.9 & 0.217 & 0.249 & 0.052 & 0.215 & 101.1 \\
        \quad + FT w/ explanations & 0.240 & 0.322 & 0.104 & 0.289 & 67.4 & 0.242 & 0.280 & 0.069 & 0.247 & 80.9 \\ 
        \midrule
        Improvement & 7.7\% & 12.0\% & 40.5\% & 11.0\% & -13.8\% & 5.6\% & 13.4\% & 20.9\% & 13.4\% & -26.5\% \\
        \midrule
        llava-v1.5-13b & 0.223 & 0.293 & 0.079 & 0.259 & 71.0 & 0.239 & 0.247 & 0.059 & 0.216 & 111.5 \\ 
        \quad + FT & 0.207 & 0.282 & 0.068 & 0.252 & 68.7 & 0.213 & 0.246 & 0.050 & 0.218 & 97.3 \\
        \quad + FT w/ explanations & 0.248 & 0.323 & 0.103 & 0.294 & 68.1 & 0.239 & 0.278 & 0.066 & 0.244 & 80.8 \\ 
        \midrule
        Improvement & 11.0\% & 10.2\% & 30.5\% & 13.5\% & -4.1\% & 0.0\% & 12.7\% & 11.6\% & 13.1\% & -27.5\% \\      
        \bottomrule
    \end{tabular}
    \vspace{-2mm}
    \caption{Results on the image description (ID) and social context description (SCD) tasks. We report METEOR (M), ROUGE-1 (R-1), ROUGE-2 (R-2), ROUGE-L (R-L), and the length of responses (Len), calculated as the number of words in the responses. ``FT'' represents fine-tuning with the ground-truth, and ``FT w/ explanations'' represents fine-tuning with both the ground-truth and the explanations. The Improvement row indicates performance gain for the FT w/ explanations setting w.r.t. zero-shot baselines. LLaVA-v1.5-7B/13B consistently achieve the best performances among all MLLMs, and exhibit improved performances after fine-tuning on explanations. }
    \label{tab:text_generation}
\end{table*}

\begin{figure}[h]
    \centering
    \includegraphics[width=0.92\linewidth]{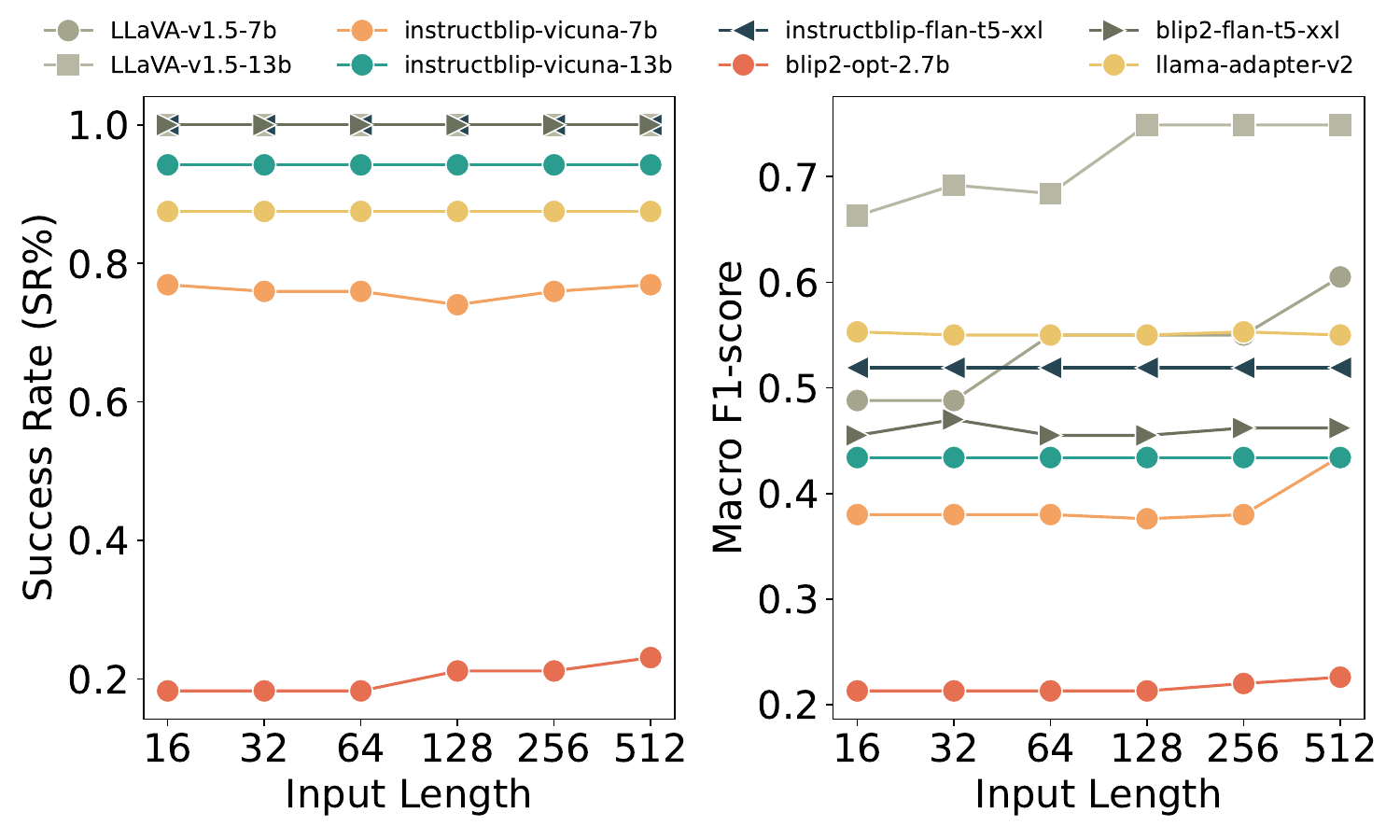}
    \vspace{-2mm}
    \caption{Success Rate (left) and macro-F1 scores (right) of varying input lengths on PolitiFact. The instruction following abilities of MLLMs remains stable across varying input lengths, and exhibit improvements as model size increases.}
    \label{fig:success_rate}
    \vspace{-3mm}
\end{figure}

\begin{table}[!t]
    \centering
    \small
    \setlength{\tabcolsep}{1mm}
    \begin{tabular}{l|ccccc}
        \toprule
        Model & Pre & Rec & F1 & Jacc & $\mathcal{L}_{\mathrm{Ham}}$ $\downarrow$ \\ 
        \midrule 
        instructblip-vicuna-13b & 0.044 & 0.230 & 0.050 & 0.032 & 0.429 \\ 
        instructblip-flan-t5-xl & 0.045 & 0.326 & 0.057 & 0.036 & 0.500 \\
        instructblip-flan-t5-xxl & 0.092 & \textbf{0.376} & 0.128 & 0.078 & \underline{0.161} \\ 
        blip2-opt-2.7b & 0.027 & 0.037 & 0.022 & 0.013 & 0.223 \\ 
        blip2-flan-t5-xl & \textbf{0.196} & 0.191 & \underline{0.157} & \underline{0.112} & 0.092 \\
        blip2-flan-t5-xxl & \underline{0.176} & 0.350 & \textbf{0.188} & \textbf{0.122} & 0.085 \\ 
        llama-adapter-v2 & 0.028 & 0.029 & 0.021 & 0.012 & \textbf{0.137} \\ 
        \midrule
        llava-v1.5-7b & 0.048 & 0.345 & 0.068 & 0.041 & 0.406 \\
        \quad + FT & 0.162 & 0.373 & 0.209 & 0.148 & 0.063  \\
        \quad + FT w/ explanations & 0.562 & 0.491 & 0.494 & 0.400 & 0.027 \\
        \midrule
        llava-v1.5-13b & 0.052 & \underline{0.361} & 0.071 & 0.043 & 0.342 \\
        \quad + FT & 0.123 & 0.441 & 0.167 & 0.113 & 0.104 \\
        \quad + FT w/ explanations & 0.533 & 0.473 & 0.474 & 0.387 & 0.027 \\
        \bottomrule
    \end{tabular}
    \caption{Results of tagging on the YouTube dataset. A ``$\downarrow$'' next to the metric indicates that lower values represent better performances. instructblip-vicuna-7b fails to produce valid predictions in this context.}
    \label{tab:tagging}
    \vspace{-4mm}
\end{table}

\noindent \textbf{Larger models exhibit better instruction-following abilities.} 
To quantify an LLM's adherence to predefined content constraints, we leverage a success rate metric, defined as the percentage of responses from a model that aligns with the requested formats. 
We see a compelling positive correlation between the parameter size of the text encoder and its ability to follow instructions and precisely classify news content. Table~\ref{tab:misinfo} shows that the macro F1-score on PolitiFact for InstructBLIP increases from 0.376 to 0.434 when the text encoder changes from Vicuna-7B to Vicuna-13B, and improves from 0.418 to 0.519 when changing from FlanT5-XL to FlanT5-XXL. 
This correlation indicates that models with larger parameter sizes are equipped with more complex reasoning abilities and a sophisticated understanding of social knowledge, which are essential components for accurately evaluating the veracity of news articles.


\begin{figure}[t]
    \centering
    \includegraphics[width=0.95\linewidth]{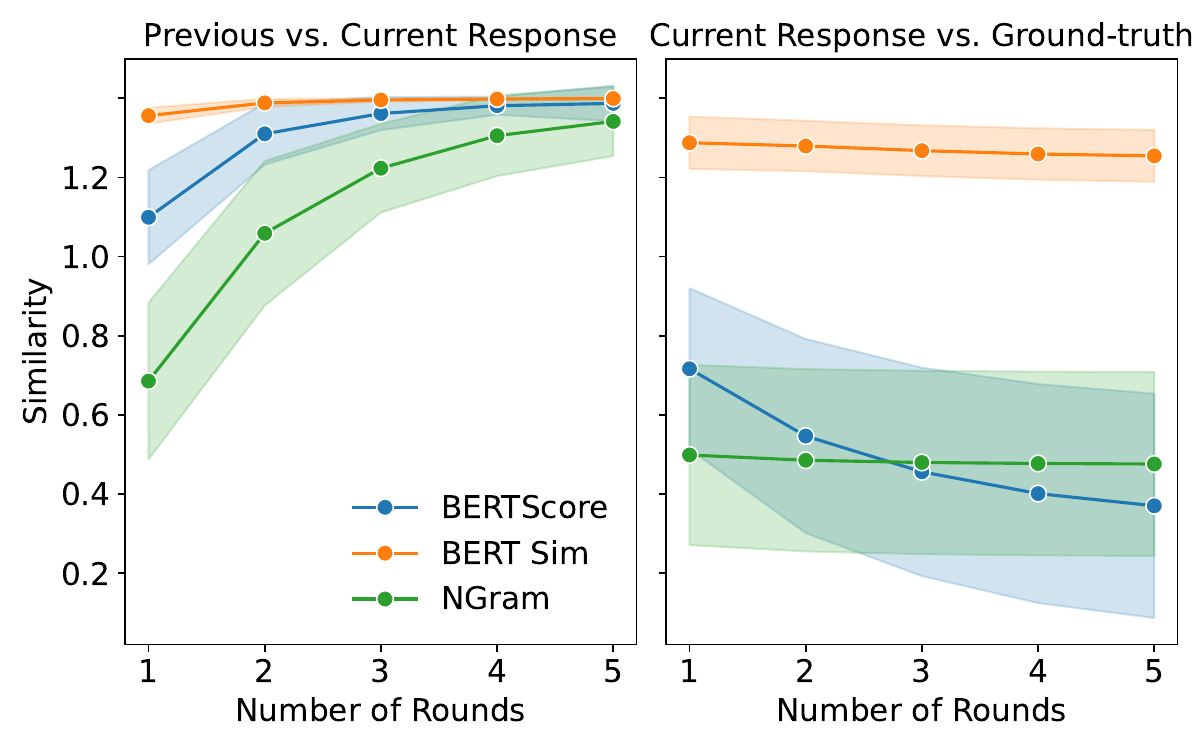}
    \vspace{-2mm}
    \caption{Left: Pairwise similarity between responses at adjacent rounds; right: similarity between response of each round and the ground-truth. }
    \label{fig:answer_similarity}
    \vspace{-2mm}
\end{figure}

Online content ranges from concise and engaging social media posts and microblogs to detailed and extensive narratives found in news articles and in-depth blog posts. This diversity in content length poses a significant challenge for MLLMs, as it requires the models to maintain their generative capabilities over varying context sizes and a wide range of information densities~\cite{peng2023yarn, peysakhovich2023attention}. To address these concerns, we vary the number of tokens used as input to detect misinformation on the PolitiFact dataset from 16 to 512 tokens. The results, as depicted in Figure \ref{fig:success_rate}, provide compelling evidence of the MLLMs' stable instruction-following abilities. Notably, we observed an increase in the macro-F1 score as the input length expanded, suggesting that MLLMs are able to leverage evidence from longer contexts for enhanced reasoning and performances. 

\section{Illustrative Uses of \benchmark}
\label{sec:case}
The \benchmark benchmark can be used to experiment with new methods for enhancing MLLMs in solving multimodal reasoning and generation tasks. We conduct two case studies, proposing new directions for strengthening MLLM capabilities.
\begin{figure}[!t]
    \centering
    \includegraphics[width=0.96\linewidth]{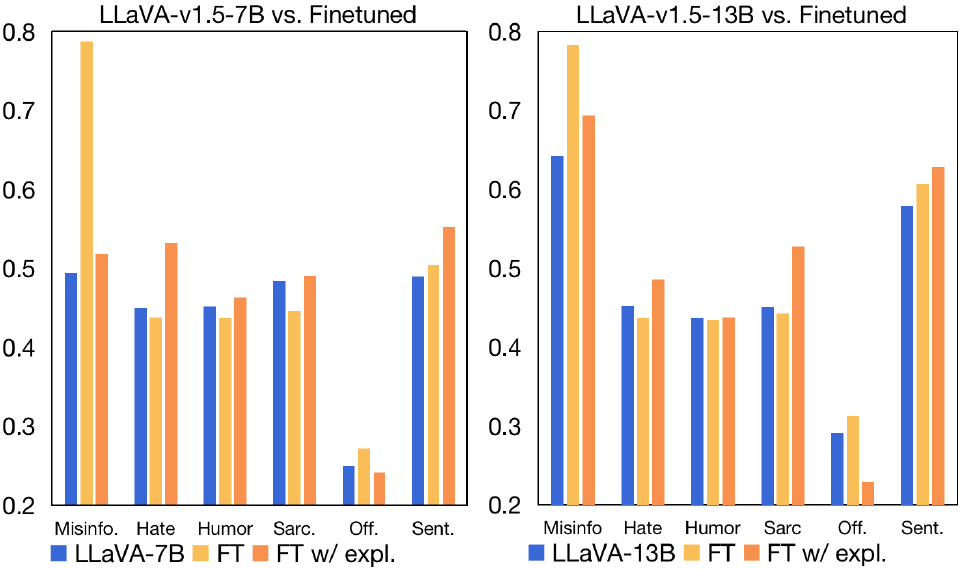}
    \vspace{-1mm}
    \caption{Results of finetuned LLaVA-v1.5-7/13B. Compared to the zero-shot baseline, finetuning with explanations (FT w/ Expl.) and standard finetuning (FT) improves performance across different sets of tasks.}
    \label{fig:finetunedLLaVA}
    \vspace{-5mm}
\end{figure}


\subsection{Can MLLMs Self-improve Its Answers?}
The ability of MLLMs to self-improve -- enhancing their answers iteratively without external supervision -- can help generate increasingly consistent and robust answers, reducing the need for human oversight.
Using our benchmark, we investigate the self-improvement capabilities of MLLMs.
The initial phase involves the model generating an answer for each question. Subsequent iterations, starting from the second round, require the model to produce new answers conditioned on the multimodal inputs and its prior responses. The iterative process is performed for six rounds. To quantitatively assess the evolution of answers across these iterations, we employed three established similarity metrics: BERTScore~\cite{zhang2019bertscore}, sentence embeddings similarity~\cite{reimers2019sentence}, and bigram similarity~\cite{kondrak2005n}. These metrics enabled us to measure the consistency of answers from one round to the next, as well as their fidelity to the ground truth.

Figure~\ref{fig:answer_similarity} displays a notable trend towards convergence in the model's answers with each iteration. For instance, the average BERTScore between answers from consecutive rounds (first to second, and second to third) exhibited a significant increase, from 0.699 to 0.910. Meanwhile, over 55\% of all answer pairs between the second and third rounds achieved a sentence embedding similarity score exceeding 0.99. 
Despite improvements in internal consistency, our analysis revealed a gradual divergence from the ground truth over successive iterations. This was evidenced by a decrease in sentence embedding similarity between MLLM-generated answers and the ground-truth (0.887 $\rightarrow$ 0.854), signaling a potential limitation in the model's ability to maintain factual accuracy in iterative generation. 

\subsection{Does finetuning MLLMs Improve Overall Performance?}
We examine whether MLLMs can improve on \benchmark via additional fine-tuning steps. Instead of fine-tuning models on separate tasks, we use the data across all different tasks at once for training and examine whether this setting still can contribute towards improvements for each task.

We employed two distinct strategies for fine-tuning. The first approach directly fine-tunes the model using the default input and output data, analogous to a standard fine-tuning setting. In the second approach, we leverage GPT-4V as a strong teacher to generate explanations after each ground truth answer for each sample. Along with the original input data, the GPT-generated explanations are augmented as additional training data.

Figure~\ref{fig:finetunedLLaVA} shows the performances of fine-tuned LLaVA-7B and 13B models along with baselines. 
With standard fine-tuning, we observe notable gains in detecting misinformation, offensiveness, and sentiment, but also drops in hate, humor, and sarcasm detection. Meanwhile, fine-tuning with explanations improved performance across a broader spectrum of tasks, e.g., increases of 18.2\% in hate speech detection and 12.7\% in sentiment analysis. Notably, text generation tasks such as image description and social context demonstrated greater gains. 
Table~\ref{tab:text_generation} further reinforces the positive effects of finetuning with explanations for text generation tasks. Compared to the zero-shot baseline, both the 7B \& 13B LLaVA models achieve higher ROUGE-2 scores on image description (40.5\% for 7B and 30.5\% for 13B). Similarly, for social context description, we observe improvements of 20.9\% and 11.6\% respectively. These improvements are accompanied by a reduction in response verbosity, highlighting the importance of explanations and rationales for improving multimodal text generation tasks. Interestingly, finetuning without explanations performs \textit{worse} than the baseline, indicating that the standard finetuning approach may not be sufficient to learn the tasks in \benchmark and signaling the need for refined finetuning strategies.

\section{Related Works}
\label{sec:related}

\noindent\textbf{Multimodal Large Language Models}: Multimodal Large Language Models (MLLMs) have demonstrated exceptional natural language understanding and generation abilities by integrating visual information with textual inputs~\cite{awadalla2023openflamingo, yu2023mm, liu2023mmc, yang2024can, li2023zone, zeng2023ipdreamer, xie2024can, xiong2024large}. 
Models such as LLaVA~\cite{liu2023improved, liu2023visual}, BLIP2~\cite{li2023blip}, InstructBLIP~\cite{dai2023instructblip}, and LLaMA-Adapter~\cite{zhang2023llama, gao2023llama} have showcased their superior performance in a range of applications. 
The success of MLLMs suggests their potential for widespread use in scenarios requiring not only factual analysis and comprehension but also subjective judgment and decision-making based on a nuanced understanding of social contexts and human perceptions. 
Our study reveals that current MLLMs still fall short of fully grasping and responding to complex social scenarios with the required depth of understanding and sensitivity. 


\vspace{0.03in}
\noindent\textbf{Benchmarking Large Language Models}:
The evaluation of LLMs is crucial for uncovering their capabilities and identifying potential risks associated with their deployment in sensitive domains~\cite{wang2024candle, wang2024mementos, wang2023abspyramid, liu2020visual, zhang2023foundation, zhao2023competeai, zong2023tilfa, xiao2023large, chan2023chatgpt}. Benchmarking efforts across various domains such as legal~\cite{deroy2023ready}, healthcare~\cite{jin2023better}, finance~\cite{zhou2023exploring}, psychology~\cite{li2023good,dan2024multiple} have provided valuable insights into LLMs such as their reliability~\cite{shu2023you}, robustness~\cite{zhu2023promptbench}, and ethical implications~\cite{sun2023aligning}. 
Despite these efforts, there remains a notable gap in the development of comprehensive multimodal benchmarks for social domains. 
In this work, we create a holistic multimodal benchmark that captures the broad spectrum of social language and interactions.

\vspace{-1mm}
\section{Conclusion}
\label{sec:conclusion}
\vspace{-1mm}
Our study presents a comprehensive evaluation of $4$ leading MLLMs on $10$ carefully constructed multimodal social media tasks from diverse domains such as misinformation, hate speech, memes, and a novel YouTube dataset, which comprises our proposed \benchmark benchmark. 
Our evaluation of the current capabilities presents the following insights: (i) zero-shot capabilities of certain MLLMs are near-random and underperform drastically in comparison to smaller fully fine-tuned models,  (ii) LLaVA-v1.5 is currently the most competitive open-source MLLM, and (iii) instruction following capabilities of MLLMs improve with their size. \benchmark also enables quantitative case studies, two of which were illustrated in this work and revealed (a) the limitations of MLLMs in self-improving their accuracy and (b) the effectiveness of fine-tuning MLLMs with labeled data. 
As benchmarks highlight current limitations and guide future research, we intend to expand \benchmark's coverage to more models and social media tasks to encourage reliable applicability of MLLMs in online spheres. 


\section{Limitations}
\label{sec:limitation}
We describe limitations of the current study settings and discuss potential directions for future works.

\subsection{Exclusion of Proprietary Models} 
This study does not focus on proprietary models like GPT-4V and Gemini for specific reasons. 
First, this research aims to spotlight the constraints of \emph{open-source MLLMs} in tackling multimodal tasks derived from social media contexts. 
This emphasis on open-source models is driven by our commitment to enhancing privacy protection.
Unlike proprietary models that aggregate data of multiple platforms onto a central server, posing significant privacy risks and operational costs, open-source models are able to process data in a decentralized way~\cite{fan2023fate, zhang2023fedpetuning}. 
This distinction not only ensures better privacy safeguards but also resonates with our objective to spotlight and scrutinize the limitations inherent within open-source frameworks when deployed in complex, real-world scenarios like social media. 
By doing so, we hope that the research community can dedicate resources towards the development of more sophisticated open-source models that address these gaps, promoting the ethos of open science. Second, proprietary models like Gemini reject images containing people and prompts associated with misinformation and hate speech. These restrictions present significant barriers to a comprehensive analysis of MLLMs' performance in handling the diverse and often complex content found on social media platforms.

\subsection{Scope of Datasets Included in Benchmark}

Online platforms facilitate several well-being discussions and provide support to potentially vulnerable members of the community~\cite{alghowinem2016multimodal, sindoni2020youcantalk}. While our current datasets consider applications of MLLMs for some safety-critical tasks like misinformation and hate detection, extensions of \benchmark should include datasets and tasks that cover applications that promote inclusivity and support-offering on online platforms. The current version of the benchmark is not ``open-world, universal, and neutral,'' the likes of which have been contested to exist~\cite{raji2021ai}, but an evolving-effort to contextualize the progress in MLLMs with respect to widely-used social media tasks.




\section{Ethical Considerations and Broader Impacts}
MLLMs are recognized for exhibiting decision-making biases, a direct consequence of biases present within their training datasets. These include but are not limited to, biases in core sociodemographic categories such as gender, race, and religion~\cite{janghorbani2023multi,ruggeri2023multi}. 
This can cause severe issues during downstream applications of MLLMs, particularly in contexts where decisions can significantly affect individual choices. 

A significant portion of the biases in MLLMs may be attributed to the data it is trained on. The annotation of subjective tasks in NLP benchmarks also requires consideration, as highlighted in various studies~\cite{aroyo2015truth,waseem2016you,alkuwatly2020identifying}. The interpretation of humor or offensive content can significantly vary across different cultural and societal backgrounds, and thus benchmarks should incorporate a broader spectrum of human viewpoints. This is also applicable to certain tasks within our benchmark, where the labels of our questions are reflective of the viewpoints of a hypothetical "average Twitter user." We recognize the importance of this diversity and inclusivity. Our hope is for subsequent research leveraging our benchmark to hopefully develop and include datasets that are more representative of social diversity and inclusiveness, thereby addressing these disparities.

One consistent theme throughout our empirical investigations is that the current performances of MLLMs in general are suboptimal. Notably, certain zero-shot MLLMs exhibit lower accuracy compared to both LLMs fine-tuned exclusively on textual data and even random scores. This underperformance is likely attributable to the insufficient training of MLLMs on tasks requiring subjective judgment and comprehension of social context. For MLLMs to achieve broader and more reliable applicability, future versions should be trained on more tasks that cover ethical, social, and cultural dimensions, thereby ensuring a more comprehensive understanding and interaction capability in diverse contexts.




\vspace{-2mm}

\section*{Acknowledgements}

\vspace{-2mm}

This research/material is based upon work supported in part by NSF grants CNS-2154118, ITE-2137724, ITE-2230692, CNS2239879, Defense Advanced Research Projects Agency (DARPA) under Agreement No. HR00112290102 (subcontract No. PO70745), CDC, and funding from Microsoft. Any opinions, findings, and conclusions or recommendations expressed in this material are those of the author(s) and do not necessarily reflect the position or policy of DARPA, DoD, SRI International, CDC, NSF, and no official endorsement should be inferred. We thank members of the CLAWS Lab for their helpful feedback.


\bibliography{cite}

\begin{thebibliography}{82}
\expandafter\ifx\csname natexlab\endcsname\relax\def\natexlab#1{#1}\fi

\bibitem[{Al~Kuwatly et~al.(2020)Al~Kuwatly, Wich, and Groh}]{alkuwatly2020identifying}
Hala Al~Kuwatly, Maximilian Wich, and Georg Groh. 2020.
\newblock \href {https://doi.org/10.18653/v1/2020.alw-1.21} {Identifying and measuring annotator bias based on annotators{'} demographic characteristics}.
\newblock In \emph{Proceedings of the Fourth Workshop on Online Abuse and Harms}, pages 184--190, Online. Association for Computational Linguistics.

\bibitem[{Alghowinem et~al.(2016)Alghowinem, Goecke, Wagner, Epps, Hyett, Parker, and Breakspear}]{alghowinem2016multimodal}
Sharifa Alghowinem, Roland Goecke, Michael Wagner, Julien Epps, Matthew Hyett, Gordon Parker, and Michael Breakspear. 2016.
\newblock Multimodal depression detection: fusion analysis of paralinguistic, head pose and eye gaze behaviors.
\newblock \emph{IEEE Transactions on Affective Computing}, 9(4):478--490.

\bibitem[{Aroyo and Welty(2015)}]{aroyo2015truth}
Lora Aroyo and Chris Welty. 2015.
\newblock Truth is a lie: Crowd truth and the seven myths of human annotation.
\newblock \emph{AI Magazine}, 36(1):15--24.

\bibitem[{Awadalla et~al.(2023)Awadalla, Gao, Gardner, Hessel, Hanafy, Zhu, Marathe, Bitton, Gadre, Sagawa et~al.}]{awadalla2023openflamingo}
Anas Awadalla, Irena Gao, Josh Gardner, Jack Hessel, Yusuf Hanafy, Wanrong Zhu, Kalyani Marathe, Yonatan Bitton, Samir Gadre, Shiori Sagawa, et~al. 2023.
\newblock Openflamingo: An open-source framework for training large autoregressive vision-language models.
\newblock \emph{arXiv:2308.01390}.

\bibitem[{Chan et~al.(2023)Chan, Cheng, Wang, Jiang, Fang, Liu, and Song}]{chan2023chatgpt}
Chunkit Chan, Jiayang Cheng, Weiqi Wang, Yuxin Jiang, Tianqing Fang, Xin Liu, and Yangqiu Song. 2023.
\newblock Chatgpt evaluation on sentence level relations: A focus on temporal, causal, and discourse relations.
\newblock \emph{arXiv:2304.14827}.

\bibitem[{Chen and Shu(2023{\natexlab{a}})}]{chen2023can}
Canyu Chen and Kai Shu. 2023{\natexlab{a}}.
\newblock Can llm-generated misinformation be detected?
\newblock \emph{arXiv:2309.13788}.

\bibitem[{Chen and Shu(2023{\natexlab{b}})}]{chen2023combating}
Canyu Chen and Kai Shu. 2023{\natexlab{b}}.
\newblock Combating misinformation in the age of llms: Opportunities and challenges.
\newblock \emph{arXiv:2311.05656}.

\bibitem[{Choi et~al.(2023)Choi, Pei, Kumar, Shu, and Jurgens}]{choi2023llms}
Minje Choi, Jiaxin Pei, Sagar Kumar, Chang Shu, and David Jurgens. 2023.
\newblock Do llms understand social knowledge? evaluating the sociability of large language models with socket benchmark.
\newblock In \emph{EMNLP 2023}.

\bibitem[{Dai et~al.(2023)Dai, Li, Li, Tiong, Zhao, Wang, Li, Fung, and Hoi}]{dai2023instructblip}
W~Dai, J~Li, D~Li, AMH Tiong, J~Zhao, W~Wang, B~Li, P~Fung, and S~Hoi. 2023.
\newblock Instructblip: Towards general-purpose vision-language models with instruction tuning. arxiv 2023.
\newblock \emph{arXiv:2305.06500}.

\bibitem[{Dan et~al.(2024)Dan, Yan, Tan, Zhou, and Lu}]{dan2024multiple}
Han-Cheng Dan, Peng Yan, Jiawei Tan, Yinchao Zhou, and Bingjie Lu. 2024.
\newblock Multiple distresses detection for asphalt pavement using improved you only look once algorithm based on convolutional neural network.
\newblock \emph{International Journal of Pavement Engineering}, 25(1):2308169.

\bibitem[{Deroy et~al.(2023)Deroy, Ghosh, and Ghosh}]{deroy2023ready}
Aniket Deroy, Kripabandhu Ghosh, and Saptarshi Ghosh. 2023.
\newblock How ready are pre-trained abstractive models and llms for legal case judgement summarization?
\newblock \emph{arXiv:2306.01248}.

\bibitem[{Fan et~al.(2023)Fan, Kang, Ma, Chen, Wei, Fan, and Yang}]{fan2023fate}
Tao Fan, Yan Kang, Guoqiang Ma, Weijing Chen, Wenbin Wei, Lixin Fan, and Qiang Yang. 2023.
\newblock Fate-llm: A industrial grade federated learning framework for large language models.
\newblock \emph{arXiv:2310.10049}.

\bibitem[{Ferrara(2020)}]{ferrara2020dynamics}
Emilio Ferrara. 2020.
\newblock \href {https://doi.org/10.1093/oxfordhb/9780190460518.013.21} {{Dynamics of Attention and Public Opinion in Social Media}}.
\newblock In \emph{{The Oxford Handbook of Networked Communication}}. Oxford University Press.

\bibitem[{Gao et~al.(2023)Gao, Han, Zhang, Lin, Geng, Zhou, Zhang, Lu, He, Yue et~al.}]{gao2023llama}
Peng Gao, Jiaming Han, Renrui Zhang, Ziyi Lin, Shijie Geng, Aojun Zhou, Wei Zhang, Pan Lu, Conghui He, Xiangyu Yue, et~al. 2023.
\newblock Llama-adapter v2: Parameter-efficient visual instruction model.
\newblock \emph{arXiv:2304.15010}.

\bibitem[{Gilardi et~al.(2023)Gilardi, Alizadeh, and Kubli}]{gilardi2023chatgpt}
Fabrizio Gilardi, Meysam Alizadeh, and Ma{\"e}l Kubli. 2023.
\newblock Chatgpt outperforms crowd-workers for text-annotation tasks.
\newblock \emph{arXiv:2303.15056}.

\bibitem[{He et~al.(2023)He, Hu, Lee, Oh, Verma, and Kumar}]{he2023survey}
Bing He, Yibo Hu, Yeon-Chang Lee, Soyoung Oh, Gaurav Verma, and Srijan Kumar. 2023.
\newblock A survey on the role of crowds in combating online misinformation: Annotators, evaluators, and creators.
\newblock \emph{arXiv:2310.02095}.

\bibitem[{He et~al.(2021)He, Ziems, Soni, Ramakrishnan, Yang, and Kumar}]{he2021racism}
Bing He, Caleb Ziems, Sandeep Soni, Naren Ramakrishnan, Diyi Yang, and Srijan Kumar. 2021.
\newblock Racism is a virus: Anti-asian hate and counterspeech in social media during the covid-19 crisis.
\newblock In \emph{ASONAM}, pages 90--94.

\bibitem[{He et~al.(2020)He, Liu, Gao, and Chen}]{he2020deberta}
Pengcheng He, Xiaodong Liu, Jianfeng Gao, and Weizhu Chen. 2020.
\newblock Deberta: Decoding-enhanced bert with disentangled attention.
\newblock In \emph{ICLR}.

\bibitem[{Holtzman et~al.(2019)Holtzman, Buys, Du, Forbes, and Choi}]{holtzman2019curious}
Ari Holtzman, Jan Buys, Li~Du, Maxwell Forbes, and Yejin Choi. 2019.
\newblock The curious case of neural text degeneration.
\newblock In \emph{ICLR}.

\bibitem[{Huang et~al.(2024)Huang, Cui, Wang, Yang, Liao, Song, Yao, and Su}]{huang2024mitigating}
Jianheng Huang, Leyang Cui, Ante Wang, Chengyi Yang, Xinting Liao, Linfeng Song, Junfeng Yao, and Jinsong Su. 2024.
\newblock Mitigating catastrophic forgetting in large language models with self-synthesized rehearsal.
\newblock \emph{arXiv:2403.01244}.

\bibitem[{Jacobi(2014)}]{jacobi2014perceptions}
Lora~L Jacobi. 2014.
\newblock Perceptions of profanity: How race, gender, and expletive choice affect perceived offensiveness.
\newblock \emph{North American Journal of Psychology}, 16(2).

\bibitem[{Janghorbani and De~Melo(2023)}]{janghorbani2023multi}
Sepehr Janghorbani and Gerard De~Melo. 2023.
\newblock \href {https://doi.org/10.18653/v1/2023.eacl-main.126} {Multi-modal bias: Introducing a framework for stereotypical bias assessment beyond gender and race in vision{--}language models}.
\newblock In \emph{EACL}, pages 1725--1735, Dubrovnik, Croatia. Association for Computational Linguistics.

\bibitem[{Jin et~al.(2023)Jin, Chandra, Verma, Hu, De~Choudhury, and Kumar}]{jin2023better}
Yiqiao Jin, Mohit Chandra, Gaurav Verma, Yibo Hu, Munmun De~Choudhury, and Srijan Kumar. 2023.
\newblock Better to ask in english: Cross-lingual evaluation of large language models for healthcare queries.
\newblock \emph{arXiv e-prints}, pages arXiv--2310.

\bibitem[{Jin et~al.(2022)Jin, Wang, Yang, Sun, Wang, Liao, and Xie}]{jin2022towards}
Yiqiao Jin, Xiting Wang, Ruichao Yang, Yizhou Sun, Wei Wang, Hao Liao, and Xing Xie. 2022.
\newblock Towards fine-grained reasoning for fake news detection.
\newblock In \emph{AAAI}, volume~36, pages 5746--5754.

\bibitem[{Kenton and Toutanova(2019)}]{kenton2019bert}
Jacob Devlin Ming-Wei~Chang Kenton and Lee~Kristina Toutanova. 2019.
\newblock Bert: Pre-training of deep bidirectional transformers for language understanding.
\newblock In \emph{NAACL}, pages 4171--4186.

\bibitem[{Kiela et~al.(2020)Kiela, Firooz, Mohan, Goswami, Singh, Ringshia, and Testuggine}]{kiela2020hateful}
Douwe Kiela, Hamed Firooz, Aravind Mohan, Vedanuj Goswami, Amanpreet Singh, Pratik Ringshia, and Davide Testuggine. 2020.
\newblock The hateful memes challenge: Detecting hate speech in multimodal memes.
\newblock \emph{NeurIPS}, 33:2611--2624.

\bibitem[{Kondrak(2005)}]{kondrak2005n}
Grzegorz Kondrak. 2005.
\newblock N-gram similarity and distance.
\newblock In \emph{International symposium on string processing and information retrieval}, pages 115--126. Springer.

\bibitem[{Lavie et~al.(2004)Lavie, Sagae, and Jayaraman}]{lavie2004significance}
Alon Lavie, Kenji Sagae, and Shyamsundar Jayaraman. 2004.
\newblock The significance of recall in automatic metrics for mt evaluation.
\newblock In \emph{AMTA}, pages 134--143. Springer.

\bibitem[{Levenshtein et~al.(1966)}]{levenshtein1966binary}
Vladimir~I Levenshtein et~al. 1966.
\newblock Binary codes capable of correcting deletions, insertions, and reversals.
\newblock In \emph{Soviet physics doklady}, volume~10, pages 707--710. Soviet Union.

\bibitem[{Li et~al.(2023{\natexlab{a}})Li, Wang, Zhang, Zhu, Wang, Hou, Lian, Luo, Yang, and Xie}]{li2023good}
Cheng Li, Jindong Wang, Yixuan Zhang, Kaijie Zhu, Xinyi Wang, Wenxin Hou, Jianxun Lian, Fang Luo, Qiang Yang, and Xing Xie. 2023{\natexlab{a}}.
\newblock The good, the bad, and why: Unveiling emotions in generative ai.
\newblock \emph{arXiv:2312.11111}.

\bibitem[{Li et~al.(2023{\natexlab{b}})Li, Li, Savarese, and Hoi}]{li2023blip}
Junnan Li, Dongxu Li, Silvio Savarese, and Steven Hoi. 2023{\natexlab{b}}.
\newblock Blip-2: Bootstrapping language-image pre-training with frozen image encoders and large language models.
\newblock \emph{arXiv:2301.12597}.

\bibitem[{Li et~al.(2023{\natexlab{c}})Li, Zeng, Feng, Gao, Liu, Liu, Lin, Tang, Hu, Liu et~al.}]{li2023zone}
Shanglin Li, Bohan Zeng, Yutang Feng, Sicheng Gao, Xuhui Liu, Jiaming Liu, Li~Lin, Xu~Tang, Yao Hu, Jianzhuang Liu, et~al. 2023{\natexlab{c}}.
\newblock Zone: Zero-shot instruction-guided local editing.
\newblock \emph{arXiv:2312.16794}.

\bibitem[{Lin(2004)}]{lin2004rouge}
Chin-Yew Lin. 2004.
\newblock Rouge: A package for automatic evaluation of summaries.
\newblock In \emph{Text summarization branches out}, pages 74--81.

\bibitem[{Liu et~al.(2023{\natexlab{a}})Liu, Wang, Yao, Chen, Song, Cho, Yacoob, and Yu}]{liu2023mmc}
Fuxiao Liu, Xiaoyang Wang, Wenlin Yao, Jianshu Chen, Kaiqiang Song, Sangwoo Cho, Yaser Yacoob, and Dong Yu. 2023{\natexlab{a}}.
\newblock Mmc: Advancing multimodal chart understanding with large-scale instruction tuning.
\newblock \emph{arXiv:2311.10774}.

\bibitem[{Liu et~al.(2020)Liu, Wang, Wang, and Ordonez}]{liu2020visual}
Fuxiao Liu, Yinghan Wang, Tianlu Wang, and Vicente Ordonez. 2020.
\newblock Visual news: Benchmark and challenges in news image captioning.
\newblock \emph{arXiv:2010.03743}.

\bibitem[{Liu et~al.(2023{\natexlab{b}})Liu, Li, Li, and Lee}]{liu2023improved}
Haotian Liu, Chunyuan Li, Yuheng Li, and Yong~Jae Lee. 2023{\natexlab{b}}.
\newblock Improved baselines with visual instruction tuning.
\newblock In \emph{NeurIPS 2023 Workshop on Instruction Tuning and Instruction Following}.

\bibitem[{Liu et~al.(2023{\natexlab{c}})Liu, Li, Wu, and Lee}]{liu2023visual}
Haotian Liu, Chunyuan Li, Qingyang Wu, and Yong~Jae Lee. 2023{\natexlab{c}}.
\newblock Visual instruction tuning.
\newblock \emph{arXiv:2304.08485}.

\bibitem[{Liu et~al.(2019)Liu, Ott, Goyal, Du, Joshi, Chen, Levy, Lewis, Zettlemoyer, and Stoyanov}]{liu2019roberta}
Yinhan Liu, Myle Ott, Naman Goyal, Jingfei Du, Mandar Joshi, Danqi Chen, Omer Levy, Mike Lewis, Luke Zettlemoyer, and Veselin Stoyanov. 2019.
\newblock Roberta: A robustly optimized bert pretraining approach.
\newblock \emph{arXiv:1907.11692}.

\bibitem[{Loper and Bird(2002)}]{loper2002nltk}
Edward Loper and Steven Bird. 2002.
\newblock Nltk: the natural language toolkit.
\newblock In \emph{Proceedings of the ACL-02 Workshop on Effective tools and methodologies for teaching natural language processing and computational linguistics-Volume 1}, pages 63--70.

\bibitem[{Ma et~al.(2022)Ma, Liu, Liu, and Han}]{ma2022curriculum}
Jiachen Ma, Yong Liu, Meng Liu, and Meng Han. 2022.
\newblock Curriculum contrastive learning for fake news detection.
\newblock In \emph{CIKM}, pages 4309--4313.

\bibitem[{Mondal et~al.(2017)Mondal, Silva, and Benevenuto}]{mondal2017measurement}
Mainack Mondal, Leandro~Ara{\'u}jo Silva, and Fabr{\'\i}cio Benevenuto. 2017.
\newblock A measurement study of hate speech in social media.
\newblock In \emph{ACM Hypertext}, pages 85--94.

\bibitem[{Papineni et~al.(2002)Papineni, Roukos, Ward, and Zhu}]{papineni2002bleu}
Kishore Papineni, Salim Roukos, Todd Ward, and Wei-Jing Zhu. 2002.
\newblock Bleu: a method for automatic evaluation of machine translation.
\newblock In \emph{ACL}, pages 311--318.

\bibitem[{Peng et~al.(2023)Peng, Quesnelle, Fan, and Shippole}]{peng2023yarn}
Bowen Peng, Jeffrey Quesnelle, Honglu Fan, and Enrico Shippole. 2023.
\newblock Yarn: Efficient context window extension of large language models.
\newblock \emph{arXiv:2309.00071}.

\bibitem[{Peysakhovich and Lerer(2023)}]{peysakhovich2023attention}
Alexander Peysakhovich and Adam Lerer. 2023.
\newblock Attention sorting combats recency bias in long context language models.
\newblock \emph{arXiv:2310.01427}.

\bibitem[{Raji et~al.(2021)Raji, Denton, Bender, Hanna, and Paullada}]{raji2021ai}
Inioluwa~Deborah Raji, Emily Denton, Emily~M Bender, Alex Hanna, and Amandalynne Paullada. 2021.
\newblock Ai and the everything in the whole wide world benchmark.
\newblock In \emph{Thirty-fifth Conference on Neural Information Processing Systems Datasets and Benchmarks Track (Round 2)}.

\bibitem[{Ramos et~al.(2023)Ramos, Elliott, and Martins}]{ramos2023captioning}
Rita Ramos, Desmond Elliott, and Bruno Martins. 2023.
\newblock \href {https://doi.org/10.18653/v1/2023.eacl-main.266} {Retrieval-augmented image captioning}.
\newblock In \emph{EACL}, pages 3666--3681, Dubrovnik, Croatia. Association for Computational Linguistics.

\bibitem[{Reimers and Gurevych(2019)}]{reimers2019sentence}
Nils Reimers and Iryna Gurevych. 2019.
\newblock \href {https://doi.org/10.18653/v1/D19-1410} {Sentence-{BERT}: Sentence embeddings using {S}iamese {BERT}-networks}.
\newblock In \emph{EMNLP-IJCNLP}, pages 3982--3992, Hong Kong, China. Association for Computational Linguistics.

\bibitem[{Ruch(2010)}]{ruch2010humor}
Willibald Ruch. 2010.
\newblock \emph{The sense of humor: Explorations of a personality characteristic}, volume~3.
\newblock Walter de Gruyter.

\bibitem[{Ruggeri and Nozza(2023)}]{ruggeri2023multi}
Gabriele Ruggeri and Debora Nozza. 2023.
\newblock \href {https://doi.org/10.18653/v1/2023.findings-acl.403} {A multi-dimensional study on bias in vision-language models}.
\newblock In \emph{Findings of the Association for Computational Linguistics: ACL 2023}, pages 6445--6455, Toronto, Canada. Association for Computational Linguistics.

\bibitem[{Savelka et~al.(2023)Savelka, Ashley, Gray, Westermann, and Xu}]{savelka2023can}
Jaromir Savelka, Kevin~D Ashley, Morgan~A Gray, Hannes Westermann, and Huihui Xu. 2023.
\newblock Can gpt-4 support analysis of textual data in tasks requiring highly specialized domain expertise?
\newblock \emph{arXiv:2306.13906}.

\bibitem[{Sharma et~al.(2020)Sharma, Bhageria, Scott, Pykl, Das, Chakraborty, Pulabaigari, and Gamb{\"a}ck}]{sharma2020semeval}
Chhavi Sharma, Deepesh Bhageria, William Scott, Srinivas Pykl, Amitava Das, Tanmoy Chakraborty, Viswanath Pulabaigari, and Bj{\"o}rn Gamb{\"a}ck. 2020.
\newblock Semeval-2020 task 8: Memotion analysis-the visuo-lingual metaphor!
\newblock In \emph{Proceedings of the Fourteenth Workshop on Semantic Evaluation}, pages 759--773.

\bibitem[{Shu et~al.(2023)Shu, Zhang, Choi, Dunagan, Card, and Jurgens}]{shu2023you}
Bangzhao Shu, Lechen Zhang, Minje Choi, Lavinia Dunagan, Dallas Card, and David Jurgens. 2023.
\newblock You don't need a personality test to know these models are unreliable: Assessing the reliability of large language models on psychometric instruments.
\newblock \emph{arXiv:2311.09718}.

\bibitem[{Shu et~al.(2020)Shu, Mahudeswaran, Wang, Lee, and Liu}]{shu2020fakenewsnet}
Kai Shu, Deepak Mahudeswaran, Suhang Wang, Dongwon Lee, and Huan Liu. 2020.
\newblock Fakenewsnet: A data repository with news content, social context, and spatiotemporal information for studying fake news on social media.
\newblock \emph{Big data}, 8(3):171--188.

\bibitem[{Sindoni(2020)}]{sindoni2020youcantalk}
Maria~Grazia Sindoni. 2020.
\newblock ‘\# youcantalk’: A multimodal discourse analysis of suicide prevention and peer support in the australian beyondblue platform.
\newblock \emph{Discourse \& Communication}, 14(2):202--221.

\bibitem[{Statista(2024)}]{YouTubeMonthlyActiveUsers}
Statista. 2024.
\newblock \href {https://www.statista.com/statistics/272014/global-social-networks-ranked-by-number-of-users/} {Most popular social networks worldwide as of january 2024, ranked by number of monthly active users}.

\bibitem[{Sun et~al.(2023)Sun, Pei, Choi, and Jurgens}]{sun2023aligning}
Huaman Sun, Jiaxin Pei, Minje Choi, and David Jurgens. 2023.
\newblock Aligning with whom? large language models have gender and racial biases in subjective nlp tasks.
\newblock \emph{arXiv:2311.09730}.

\bibitem[{Vosoughi et~al.(2018)Vosoughi, Roy, and Aral}]{vosoughi2018spread}
Soroush Vosoughi, Deb Roy, and Sinan Aral. 2018.
\newblock The spread of true and false news online.
\newblock \emph{science}, 359(6380):1146--1151.

\bibitem[{Wang et~al.(2024{\natexlab{a}})Wang, Fang, Li, Shi, Ding, Xu, Wang, Bai, Liu, Cheng et~al.}]{wang2024candle}
Weiqi Wang, Tianqing Fang, Chunyang Li, Haochen Shi, Wenxuan Ding, Baixuan Xu, Zhaowei Wang, Jiaxin Bai, Xin Liu, Jiayang Cheng, et~al. 2024{\natexlab{a}}.
\newblock Candle: Iterative conceptualization and instantiation distillation from large language models for commonsense reasoning.
\newblock \emph{arXiv:2401.07286}.

\bibitem[{Wang et~al.(2020)Wang, Wei, Dong, Bao, Yang, and Zhou}]{wang2020minilm}
Wenhui Wang, Furu Wei, Li~Dong, Hangbo Bao, Nan Yang, and Ming Zhou. 2020.
\newblock Minilm: Deep self-attention distillation for task-agnostic compression of pre-trained transformers.
\newblock \emph{NeurIPS}, 33:5776--5788.

\bibitem[{Wang et~al.(2024{\natexlab{b}})Wang, Zhou, Liu, Lu, Xu, He, Yoon, Lu, Bertasius, Bansal et~al.}]{wang2024mementos}
Xiyao Wang, Yuhang Zhou, Xiaoyu Liu, Hongjin Lu, Yuancheng Xu, Feihong He, Jaehong Yoon, Taixi Lu, Gedas Bertasius, Mohit Bansal, et~al. 2024{\natexlab{b}}.
\newblock Mementos: A comprehensive benchmark for multimodal large language model reasoning over image sequences.
\newblock \emph{arXiv:2401.10529}.

\bibitem[{Wang et~al.(2023)Wang, Shi, Wang, Fang, Zhang, Choi, Liu, and Song}]{wang2023abspyramid}
Zhaowei Wang, Haochen Shi, Weiqi Wang, Tianqing Fang, Hongming Zhang, Sehyun Choi, Xin Liu, and Yangqiu Song. 2023.
\newblock Abspyramid: Benchmarking the abstraction ability of language models with a unified entailment graph.
\newblock \emph{arXiv:2311.09174}.

\bibitem[{Waseem(2016)}]{waseem2016you}
Zeerak Waseem. 2016.
\newblock Are you a racist or am i seeing things? annotator influence on hate speech detection on twitter.
\newblock In \emph{Proceedings of the first workshop on NLP and computational social science}, pages 138--142.

\bibitem[{Wolf et~al.(2020)Wolf, Debut, Sanh, Chaumond, Delangue, Moi, Cistac, Rault, Louf, Funtowicz et~al.}]{wolf2020transformers}
Thomas Wolf, Lysandre Debut, Victor Sanh, Julien Chaumond, Clement Delangue, Anthony Moi, Pierric Cistac, Tim Rault, R{\'e}mi Louf, Morgan Funtowicz, et~al. 2020.
\newblock Transformers: State-of-the-art natural language processing.
\newblock In \emph{EMNLP}, pages 38--45.

\bibitem[{Xiao et~al.(2023)Xiao, Jin, Bai, Wu, Yang, Luo, Yu, Zhao, Liu, Chen et~al.}]{xiao2023large}
Yijia Xiao, Yiqiao Jin, Yushi Bai, Yue Wu, Xianjun Yang, Xiao Luo, Wenchao Yu, Xujiang Zhao, Yanchi Liu, Haifeng Chen, et~al. 2023.
\newblock Large language models can be good privacy protection learners.
\newblock \emph{arXiv:2310.02469}.

\bibitem[{Xie et~al.(2024)Xie, Chen, Jia, Ye, Shu, Bibi, Hu, Torr, Ghanem, and Li}]{xie2024can}
Chengxing Xie, Canyu Chen, Feiran Jia, Ziyu Ye, Kai Shu, Adel Bibi, Ziniu Hu, Philip Torr, Bernard Ghanem, and Guohao Li. 2024.
\newblock Can large language model agents simulate human trust behaviors?
\newblock \emph{arXiv:2402.04559}.

\bibitem[{Xiong et~al.(2024)Xiong, Payani, Kompella, and Fekri}]{xiong2024large}
Siheng Xiong, Ali Payani, Ramana Kompella, and Faramarz Fekri. 2024.
\newblock Large language models can learn temporal reasoning.
\newblock \emph{arXiv:2401.06853}.

\bibitem[{Yang et~al.(2023)Yang, Ma, Liu, and Liu}]{yang2023multi}
Pingping Yang, Jiachen Ma, Yong Liu, and Meng Liu. 2023.
\newblock Multi-modal transformer for fake news detection.
\newblock \emph{Mathematical Biosciences and Engineering: MBE}, 20(8):14699--14717.

\bibitem[{Yang et~al.(2022)Yang, Wang, Jin, Li, Lian, and Xie}]{yang2022reinforcement}
Ruichao Yang, Xiting Wang, Yiqiao Jin, Chaozhuo Li, Jianxun Lian, and Xing Xie. 2022.
\newblock Reinforcement subgraph reasoning for fake news detection.
\newblock In \emph{KDD}, pages 2253--2262.

\bibitem[{Yang et~al.(2024)Yang, Xiong, Payani, Shareghi, and Fekri}]{yang2024can}
Yuan Yang, Siheng Xiong, Ali Payani, Ehsan Shareghi, and Faramarz Fekri. 2024.
\newblock Can llms reason in the wild with programs?
\newblock \emph{arXiv:2406.13764}.

\bibitem[{Yu et~al.(2023)Yu, Yang, Li, Wang, Lin, Liu, Wang, and Wang}]{yu2023mm}
Weihao Yu, Zhengyuan Yang, Linjie Li, Jianfeng Wang, Kevin Lin, Zicheng Liu, Xinchao Wang, and Lijuan Wang. 2023.
\newblock Mm-vet: Evaluating large multimodal models for integrated capabilities.
\newblock \emph{arXiv:2308.02490}.

\bibitem[{Zannettou et~al.(2018)Zannettou, Caulfield, Blackburn, De~Cristofaro, Sirivianos, Stringhini, and Suarez-Tangil}]{zannettou2018memes}
Savvas Zannettou, Tristan Caulfield, Jeremy Blackburn, Emiliano De~Cristofaro, Michael Sirivianos, Gianluca Stringhini, and Guillermo Suarez-Tangil. 2018.
\newblock \href {https://doi.org/10.1145/3278532.3278550} {On the origins of memes by means of fringe web communities}.
\newblock In \emph{ACM IMC}, IMC '18, page 188–202, New York, NY, USA. Association for Computing Machinery.

\bibitem[{Zeng et~al.(2023)Zeng, Li, Feng, Li, Gao, Liu, Li, Tang, Liu, and Zhang}]{zeng2023ipdreamer}
Bohan Zeng, Shanglin Li, Yutang Feng, Hong Li, Sicheng Gao, Jiaming Liu, Huaxia Li, Xu~Tang, Jianzhuang Liu, and Baochang Zhang. 2023.
\newblock Ipdreamer: Appearance-controllable 3d object generation with image prompts.
\newblock \emph{arXiv:2310.05375}.

\bibitem[{Zhai et~al.(2024)Zhai, Tong, Li, Cai, Qu, Lee, and Ma}]{zhai2024investigating}
Yuexiang Zhai, Shengbang Tong, Xiao Li, Mu~Cai, Qing Qu, Yong~Jae Lee, and Yi~Ma. 2024.
\newblock Investigating the catastrophic forgetting in multimodal large language model fine-tuning.
\newblock In \emph{Conference on Parsimony and Learning}, pages 202--227. PMLR.

\bibitem[{Zhang et~al.(2023{\natexlab{a}})Zhang, Liu, Li, Xie, Kim, and Wang}]{zhang2023foundation}
Peiyan Zhang, Haoyang Liu, Chaozhuo Li, Xing Xie, Sunghun Kim, and Haohan Wang. 2023{\natexlab{a}}.
\newblock Foundation model-oriented robustness: Robust image model evaluation with pretrained models.
\newblock \emph{arXiv:2308.10632}.

\bibitem[{Zhang et~al.(2023{\natexlab{b}})Zhang, Han, Zhou, Hu, Yan, Lu, Li, Gao, and Qiao}]{zhang2023llama}
Renrui Zhang, Jiaming Han, Aojun Zhou, Xiangfei Hu, Shilin Yan, Pan Lu, Hongsheng Li, Peng Gao, and Yu~Qiao. 2023{\natexlab{b}}.
\newblock Llama-adapter: Efficient fine-tuning of language models with zero-init attention.
\newblock \emph{arXiv:2303.16199}.

\bibitem[{Zhang et~al.(2019)Zhang, Kishore, Wu, Weinberger, and Artzi}]{zhang2019bertscore}
Tianyi Zhang, Varsha Kishore, Felix Wu, Kilian~Q Weinberger, and Yoav Artzi. 2019.
\newblock Bertscore: Evaluating text generation with bert.
\newblock In \emph{International Conference on Learning Representations}.

\bibitem[{Zhang et~al.(2023{\natexlab{c}})Zhang, Yang, Dai, Wang, Yu, Qu, and Xu}]{zhang2023fedpetuning}
Zhuo Zhang, Yuanhang Yang, Yong Dai, Qifan Wang, Yue Yu, Lizhen Qu, and Zenglin Xu. 2023{\natexlab{c}}.
\newblock Fedpetuning: When federated learning meets the parameter-efficient tuning methods of pre-trained language models.
\newblock In \emph{Findings of the Association for Computational Linguistics: ACL 2023}, pages 9963--9977.

\bibitem[{Zhao et~al.(2023)Zhao, Wang, Zhang, Jin, Zhu, Chen, and Xie}]{zhao2023competeai}
Qinlin Zhao, Jindong Wang, Yixuan Zhang, Yiqiao Jin, Kaijie Zhu, Hao Chen, and Xing Xie. 2023.
\newblock Competeai: Understanding the competition behaviors in large language model-based agents.
\newblock \emph{arXiv:2310.17512}.

\bibitem[{Zhou et~al.(2023)Zhou, Cao, Huang, Ye, Zhang, Liu, Xie, Hua, and Kim}]{zhou2023exploring}
Peilin Zhou, Meng Cao, You-Liang Huang, Qichen Ye, Peiyan Zhang, Junling Liu, Yueqi Xie, Yining Hua, and Jaeboum Kim. 2023.
\newblock Exploring recommendation capabilities of gpt-4v (ision): A preliminary case study.
\newblock \emph{arXiv:2311.04199}.

\bibitem[{Zhu et~al.(2023{\natexlab{a}})Zhu, Chen, Shen, Li, and Elhoseiny}]{zhu2023minigpt}
Deyao Zhu, Jun Chen, Xiaoqian Shen, Xiang Li, and Mohamed Elhoseiny. 2023{\natexlab{a}}.
\newblock Minigpt-4: Enhancing vision-language understanding with advanced large language models.
\newblock \emph{arXiv:2304.10592}.

\bibitem[{Zhu et~al.(2023{\natexlab{b}})Zhu, Wang, Zhou, Wang, Chen, Wang, Yang, Ye, Gong, Zhang et~al.}]{zhu2023promptbench}
Kaijie Zhu, Jindong Wang, Jiaheng Zhou, Zichen Wang, Hao Chen, Yidong Wang, Linyi Yang, Wei Ye, Neil~Zhenqiang Gong, Yue Zhang, et~al. 2023{\natexlab{b}}.
\newblock Promptbench: Towards evaluating the robustness of large language models on adversarial prompts.
\newblock \emph{arXiv:2306.04528}.

\bibitem[{Zong et~al.(2023)Zong, Wang, Xu, Zheng, Shi, Wang, Song, Wong, and See}]{zong2023tilfa}
Qing Zong, Zhaowei Wang, Baixuan Xu, Tianshi Zheng, Haochen Shi, Weiqi Wang, Yangqiu Song, Ginny~Y Wong, and Simon See. 2023.
\newblock Tilfa: A unified framework for text, image, and layout fusion in argument mining.
\newblock \emph{EMNLP 2023}, page 139.

\end{thebibliography}

\appendix


\begin{table*}[ht]
    \centering
    \small
    \begin{tabularx}{0.96\textwidth}{l|p{6.4cm}p{3.6cm}X}
    \toprule
        Dataset & Domain & Modality & Size \\ \midrule
        PolitiFact & \multirow{2}{*}{misinformation} & \multirow{2}{=}{news content, online posts, \newline images, user metadata} & 485  \\
        
        GossipCop &  &  & 12,840 \\ \midrule
        \multirow{1}{*}{Hateful Memes} & hate speech, OCR & images, embedded text & 12,143 \\ \midrule
        \multirow{1}{*}{Memotion} & sentiment, humor, OCR,  offensiveness, sarcasm & images, embedded text & \multirow{1}{*}{10,000} \\ \midrule
        YouTube & tagging & images, text, channels & 1,963,697 \\ 
        \bottomrule
    \end{tabularx}
    \caption{Statistics of the \benchmark benchmark. 
    }
    \label{tab:dataset}
\end{table*}
\begin{table*}[!ht]
    \centering
    \begin{tabularx}{\textwidth}{lXX}
    \toprule
        Task & Prompt & Tags \\ \hline
        Misinformation Detection & Is the following news misinformation? \textbf{[NEWS]} & misinformation, not misinformation \\ \hline
        Hate Speech Detection & Is the following meme hateful? & hateful, not\_hateful \\ \hline
        Humor & Is the following meme humorous? & humorous, not\_humorous \\ \hline
        Sarcasm & Is the following meme sarcastic? & sarcastic, not\_sarcastic \\ \hline
        Offensiveness & Is the following meme offensive? & offensive, not\_offensive \\ \hline
        
        Sentiment Analysis & What is the overall sentiment expressed through this meme? & positive, neutral, negative \\ \hline
        
        OCR & What is the text in the image? & / \\ \hline

        Image Description & Describe the scene, such as its major subjects, colors, and texture. & / \\ \hline
        Social Context Description & Describe the cultural and social context of the image. What particular groups is the image and text targeting at? & / \\ \hline
        Tagging & Predict the tags of the following online video given its title, description, and thumbnail image. Different tags must be separated by commas. \newline Title: \textbf{[TITLE]} \newline Description: \textbf{[DESCRIPTION]} & (See Table~\ref{tab:YouTubeTags} for the list of tags for YouTube videos) \\
        
        \bottomrule
    \end{tabularx}
    \caption{Prompts and possible values for each task.}
    \label{tab:prompt}
\end{table*}

\begin{table*}[!ht]
    \centering
    \begin{tabularx}{\textwidth}{X}
    \toprule
        \multicolumn{1}{c}{YouTube Tags} \\ \hline
        action-adventure\_game, action\_game, american\_football, association\_football, baseball, basketball, boxing, business, casual\_game, christian\_music, classical\_music, country\_music, cricket, electronic\_music, entertainment, fashion, film, food, golf, health, hip\_hop\_music, hobby, humour, ice\_hockey, independent\_music, jazz, knowledge, lifestyle, military, mixed\_martial\_arts, motorsport, music, music\_of\_asia, music\_of\_latin\_america, music\_video\_game, performing\_arts, pet, physical\_attractiveness, physical\_fitness, politics, pop\_music, professional\_wrestling, puzzle\_video\_game, racing\_video\_game, reggae, religion, rhythm\_and\_blues, rock\_music, role-playing\_video\_game, simulation\_video\_game, society, soul\_music, sport, sports\_game, strategy\_video\_game, technology, television\_program, tennis, tourism, vehicle, video\_game\_culture, volleyball \\
        \bottomrule
    \end{tabularx}
    \caption{Set of tags for YouTube videos}
    \label{tab:YouTubeTags}
\end{table*}

\begin{table*}[!ht]
    \centering
    \begin{tabular}{lll|lll}
    \toprule
        IETF & Language & Count & IETF & Language & Count \\ 
        \midrule
        en & English & 227,341 & kl & Kalaallisut & 1 \\ 
        en-GB & English (United Kingdom) & 38,669 & ia & Interlingua & 1 \\ 
        en-US & English (United States) & 9,398 & fr-CH & French as spoken in Switzerland & 1 \\ 
        ko & Korean & 7,873 & yo & Yoruba & 1 \\ 
        de & German & 3,988 & ff & Fula & 1 \\ 
        ja & Japanese & 3,843 & ba & Bashkir & 1 \\ 
        ru & Russian & 3,575 & sd & Sindhi & 1 \\ 
        fr & Franch & 3,091 & gd & Scottish Gaelic & 1 \\ 
        es & Spanish & 2,799 & as & Assamese & 1 \\ 
        pt & Portuguese & 2,622 & oc & Occitan & 1 \\ 
        \bottomrule
    \end{tabular}
    \caption{Language Distribution of YouTube videos in the \emph{YouTube2M} dataset. ``IETF'' represents IETF BCP 47 language tag, the standardized code for identifying human languages on the Internet.}
    \label{tab:language_distribution}
\end{table*}

\section{Task Selection}

\label{sec:task_selection}

The selection of tasks and datasets in \benchmark centers around three key criteria:

\begin{itemize}
    \item Tasks that require multimodal understanding of both textual and image domains;
    \item Tasks directly related to the dynamics of social media platforms;
    \item Tasks that have undergone rigorous evaluation in subsequent research, which affirms their validity as a benchmark.
\end{itemize}

The task selection process started with a comprehensive literature review through NLP conferences (ACL, EMNLP, NAACL, SemEval), Machine Learning conferences (NeurIPS and ICML) and Data Mining conferences (KDD and SIGIR) since 2019. Papers satisfying these criteria were retained. 
Our final list of tasks, while collectively categorized under multimodal engagements in social media contexts, each distinctly require a variety of cognitive capabilities. Some of these capabilities intersect across different tasks, while others are unique to specific challenges. Every task demands that models not only comprehend textual instructions but also accurately interpret relevant visual information to solve the task. 

\section{The \emph{YouTube2M} dataset}
\label{app:YouTube2M}
\subsection{Distribution of Languages}

Table~\ref{tab:language_distribution} shows the language distribution of YouTube videos in \emph{YouTube2M}. 
There are 138 unique languages in \emph{YouTube2M}. 323,007 videos have explicitly specified their default languages, representing 16.45\% of the total 1,963,697 videos. We provide a detailed breakdown of the languages, showcasing the distribution of the top 10 most and least popular languages within the dataset.  Our findings reveal a long-tail distribution in language popularity. Notably, English (including en, en-GB, en-US) dominates the dataset with 275,408 videos, accounting for 85.3\% of videos with a specified language. In contrast, the ten least common languages each only appear once. 

\begin{table}[!ht]
    \centering
    \begin{tabular}{ll}
    \toprule
        Tag & Count \\ 
        \midrule
        Music & 524,369 \\ 
        Video\_game\_culture & 475,124 \\ 
        Action\_game & 406,582 \\ 
        Lifestyle\_(sociology) & 383,092 \\ 
        Action-adventure\_game & 344,743 \\ 
        Role-playing\_video\_game & 320,837 \\ 
        Entertainment & 201,722 \\ 
        Strategy\_video\_game & 200,968 \\ 
        Society & 190,023 \\ 
        Pop\_music & 171,038 \\ 
        Rock\_music & 158,922 \\ 
    \bottomrule
    \end{tabular}
    \caption{The most popular tags in \emph{YouTube2M}}
    \label{tab:popular_tags}
\end{table}

\subsection{Distribution of Tags}

The \emph{YouTube2M} dataset encompasses a rich variety of 62 unique tags, with 1,389,219 videos bearing the top 5 tags, as shown in Table~\ref{tab:popular_tags}. Note that a video can have multiple tags. This accounts for 70.7\% of the entire dataset. We observe a strong inclination towards gaming and music content. 

\subsection{Channel Information}

\begin{table*}[!ht]
    \centering
    \begin{tabularx}{0.98\textwidth}{p{0.2\textwidth}|X|p{0.1\textwidth}}
    \toprule
        Channel & Description & \#Videos \\ 
        \midrule
        euronews & European television news network, headquartered in Brussels, Belgium & 4737 \\ 
        Al Jazeera English & 24-hour English news channel headquartered in Doha, Qatar & 4655 \\ 
        IGN & American video game media featuring videos about the latest gaming \& entertainment news and events & 4041 \\ 
        Fox News & 24-hour all-encompassing news service dedicated to delivering breaking news and political \& business news & 3504 \\ 
        WWE & World Wrestling Entertainment (WWE) original shows & 2847 \\ 
        The Young Turks & American progressive news commentary show & 2345 \\ 
        ESPN & Multimedia sports entertainment channel & 2026 \\ 
        Movieclips & Online movie clips collection & 1883 \\ 
        CNN & Multinational news channel and website & 1808 \\ 
        The Majority Report w/ Sam Seder & daily political talk show & 1655 \\ 
        \bottomrule
    \end{tabularx}
    \caption{YouTube channels with the most videos in the dataset.}
\end{table*}

There are 604,340 unique channels associated with the videos in the dataset. The most popular 10 channels and their associated videos are shown in Table~\ref{tab:popular_tags}. As observed from the statistics, a large portion of videos propagated in online social networks are centered around news and sports, signifying the popularity of these topics among online discourse.

\section{Details about Datasets}

\subsection{Tagging}

The tagging task focuses on predicting appropriate ``topic categories'' for YouTube videos, chosen from a predefined set as listed in Table~\ref{tab:YouTubeTags}. 
These topics make it easier for users to find videos that match their interests but also enhance the overall content management strategy. 
This dataset exemplifies the necessity of multimodal understanding in categorizing online video content. The dataset is licensed under the Apache 2.0 License.

Given the substantial volume of the \emph{YouTube2M} dataset, evaluation and fine-tuning on the entire dataset presents challenges such as runtime costs and catastrophic forgetting~\cite{huang2024mitigating, zhai2024investigating}, where LLMs severely forget previously acquired information upon being trained on new data. 
To address potential biases and the predominance of YouTube data in tagging tasks, we strategically curated a subset of 2,000 examples from \emph{YouTube2M}, aiming to mitigate any disproportionate influence of tagging tasks on the fine-tuning process. We partitioned the sampled dataset into training and test sets with an 80:20 ratio. 

\subsection{Misinformation datasets}
We consider two datasets under the misinformation detection theme: PolitiFact and GossipCop. Both datasets were curated by~\citet{shu2020fakenewsnet}, distributed under the CC-BY-SA License, and are publicly available for download at \url{https://github.com/KaiDMML/FakeNewsNet}/.
\subsubsection{PolitiFact}
This dataset contains news content from the fact-checking website PolitiFact\footnote{\url{https://www.politifact.com/}}, which focuses on political discourse, and contains the title, body, images, and user metadata from news articles. The dataset contains 485 news articles. Each article is annotated into one of the two categories: `fake'
 and `real.'
\subsubsection{GossipCop} 
This dataset contains news content from GossipCop, which targets the realm of entertainment news, and includes the title, body, images, from the news articles. The article contains 12,840 new articles, each of which is categorized into one of the two categories: `fake' and `real.'

\subsection{Hateful Memes}
The Hateful Memes dataset contains 12,840 memes that were designed to include meme-like visuals with text laid over them. Since a unimodal classifier (i.e., text-only or image-only) would struggle to make an inference about the hateful nature of the memes without considering both the modalities, they present a unique opportunity to test the multimodal reasoning capabilities of MLLMs. The designed memes were manually annotated to be in one of the two categories: `hateful' or `benign.' The dataset is distributed under the MIT License.

\subsection{Memotion}

The Memotion dataset comprises 12,143 memes, each meticulously annotated with labels that categorize the memes according to their sentiment (positive, negative, neutral), the type of emotion they convey (sarcastic, funny, offensive, motivational), and the intensity of the expressed emotion. 
The emotion class and the overall sentiment were manually labeled by Amazon Mechanical Turk (AMT) workers. 
The dataset is distributed under the Community Free Resource License\footnote{\url{https://www.figma.com/legal/community-free-resource-license/}}.

\section{Details about Experiments}

\subsection{Implementation Details}
\label{app:implementation_details}
\noindent \textbf{Benchmark Evaluation} 
For inference, we use Nucleus Sampling~\cite{holtzman2019curious} with a probability threshold of 0.9, a temperature of 1.0, and a maximum output length of 256 tokens. To account for the randomness in the generation process, we run each experiment with 3 random seeds and report the average results. All experiments were conducted on a server with 5 A100 80GB GPUs. 
The models are implemented using the Transformers library~\cite{wolf2020transformers}. 
We use the NLTK package~\cite{loper2002nltk} to calculate BLEU scores, the \textsf{rouge}\footnote{\url{https://github.com/pltrdy/rouge}} package to calculate ROUGE scores and the \textsf{sentence-bert}\footnote{\url{https://github.com/UKPLab/sentence-transformers}} package to calculate sentence embedding similarities, respectively.

\noindent \textbf{Model Finetuning.} We finetuned the models for 1 epoch using a batch size of 16, a warmup ratio of 0.03, a learning rate of 2e-4 and a cosine annealing learning rate scheduler. 

\subsection{Evaluation Metrics}

\textbf{Classification.} For classification tasks, we employ metrics including macro precision, macro recall, macro F1-score, accuracy (Acc), and Area Under the Curve (AUC), reflecting the comprehensive assessment of the models' tagging proficiency.

\noindent \textbf{Tagging.} For the tagging task, we additionally leverage Hamming Loss and Jaccard index. Hamming loss ($\mathcal{L}_{\mathrm{Hamming}}$) is used to measure the fraction of labels that are incorrectly predicted:
\begin{equation}
    \mathcal{L}_{\mathrm{Hamming}} = \frac{1}{N} \sum_{i=1}^{N} \frac{1}{|L|} \sum_{j=1}^{|L|} \mathrm{XOR} (y_{ij}, \hat{y}_{ij})
\end{equation}
where $y_{ij} \in \{0, 1\}$ is a binary variable that indicates whether example $i$ has label $j$.  $\hat{y}_{ij} \in \{0, 1\}$ is the predicted binary variable. $N$ is the number of examples in the dataset, and $L$ is the set of labels. 
\begin{table}[ht]
    \centering
    \begin{tabular}{lllll}
        \toprule
        Task & Acc & Pre & Rec & F1 \\ 
        \midrule
        Misinformation & 0.835 & 0.758 & 0.773 & 0.765 \\ 
        Humor & 0.547 & 0.486 & 0.483 & 0.477 \\ 
        Sarcasm & 0.717 & 0.544 & 0.519 & 0.503 \\ 
        Offensive & 0.389 & 0.403 & 0.432 & 0.364 \\ 
        Sentiment & 0.244 & 0.272 & 0.162 & 0.198 \\ 
        Hate Speech & 0.670 & 0.626 & 0.695 & 0.614 \\ 
        \bottomrule
    \end{tabular}
    \caption{Results on GPT4V. }
    \label{tab:GPT4V}
\end{table}
Jaccard index is defined as the size of the intersection between the predicted labels and the ground-truth divided by the size of their union:
\begin{equation}
    \mathrm{Jaccard} = \frac{1}{N} \sum_{i=1}^{N} \frac{|Y_i \cap \hat{Y}_i|}{|Y_i \cup \hat{Y}_i|}
\end{equation}
where $N$ is the total number of examples.
$\hat{Y}_i$ and $Y_i$ are the set of predicted and ground-truth labels for example $i$.

\noindent \textbf{OCR.} We use word error rate (WER), character error rate (CER), and BLEU scores~\cite{papineni2002bleu}. 
The word error rate (WER) and character error rate (CER) are derived from the Levenshtein distance~\cite{levenshtein1966binary}, defined as: 
\begin{eqnarray}
    \mathrm{WER} = \frac{|W_S| + |W_D| + |W_I|}{|W|} \\
    \mathrm{CER} = \frac{|C_S| + |C_D| + |C_I|}{|C|} 
\end{eqnarray}
where $|W|$ and $|C|$ are the number of words and characters in the ground-truth. $|W_S|$, $|W_D|$, and $|W_I|$ are the number of substitutions, deletions, and insertions at the word, and $|C_S|$, $|C_D|$, and $|C_I|$ are at the character level.

\label{app:fullEvalResults}

\begin{table*}[!ht]
    \centering
    \small
    \begin{tabular}{l|ccccccccc}
        \toprule
        Memotion & $\mathrm{P}_{\mathrm{macro}}$ & $\mathrm{R}_{\mathrm{macro}}$ & $\mathrm{F1}_{\mathrm{macro}}$ & WER & CER & BLEU1 & BLEU2 & BLEU3 & BLEU4 \\
        \midrule
        llava-v1.5-7b & 0.651 & 0.455 & 0.535 & 46.7 & 40.8 & 0.495 & 0.454 & 0.410 & 0.365 \\
        llava-v1.5-13b & 0.665 & 0.470 & 0.551 & 45.0 & 39.2 & 0.521 & 0.481 & 0.437 & 0.396 \\
        instructblip-flan-t5-xl & 0.850 & 0.482 & 0.615 & 46.3 & 42.2 & 0.490 & 0.449 & 0.405 & 0.363 \\
        instructblip-flan-t5-xxl & 0.808 & 0.441 & 0.571 & 50.0 & 45.3 & 0.445 & 0.406 & 0.365 & 0.326 \\
        instructblip-vicuna-7b & 0.853 & 0.558 & 0.675 & 38.7 & 35.1 & 0.569 & 0.534 & 0.497 & 0.459 \\
        instructblip-vicuna-13b & 0.834 & 0.451 & 0.585 & 48.9 & 44.9 & 0.459 & 0.425 & 0.387 & 0.350 \\
        blip2-opt-2.7b & 0.774 & 0.562 & 0.651 & 40.7 & 35.1 & 0.537 & 0.493 & 0.451 & 0.407 \\
        blip2-flan-t5-xl & 0.825 & 0.593 & 0.690 & 37.8 & 31.3 & 0.606 & 0.546 & 0.488 & 0.432 \\
        blip2-flan-t5-xxl & 0.791 & 0.623 & 0.697 & 36.3 & 27.8 & 0.632 & 0.569 & 0.507 & 0.448 \\
        llama-adapter-v2 & 0.183 & 0.084 & 0.115 & 94.5 & 82.2 & 0.059 & 0.036 & 0.027 & 0.021 \\
        \midrule
        Hateful Memes & $\mathrm{P}_{\mathrm{macro}}$ & $\mathrm{R}_{\mathrm{macro}}$ & $\mathrm{F1}_{\mathrm{macro}}$ & WER & CER & BLEU1 & BLEU2 & BLEU3 & BLEU4 \\ 
        \midrule
        LLaVA-v1.5-7b & 0.560 & 0.441 & 0.493 & 42.3 & 34.1 & 0.535 & 0.500 & 0.468 & 0.412 \\
        LLaVA-v1.5-13b & 0.619 & 0.469 & 0.534 & 40.3 & 32.8 & 0.568 & 0.536 & 0.506 & 0.450 \\
        instructblip-flan-t5-xl & 0.839 & 0.584 & 0.689 & 34.6 & 27.3 & 0.618 & 0.572 & 0.524 & 0.467 \\ 
        instructblip-flan-t5-xxl & 0.829 & 0.536 & 0.651 & 39.7 & 32.7 & 0.550 & 0.506 & 0.465 & 0.408 \\
        instructblip-vicuna-7b & 0.835 & 0.644 & 0.727 & 29.7 & 22.5 & 0.670 & 0.629 & 0.587 & 0.529 \\
        instructblip-vicuna-13b & 0.824 & 0.564 & 0.670 & 37.1 & 30.2 & 0.592 & 0.552 & 0.507 & 0.451 \\
        blip2-opt-2.7b & 0.759 & 0.653 & 0.702 & 29.4 & 21.7 & 0.646 & 0.599 & 0.551 & 0.494 \\ 
        blip2-flan-t5-xl & 0.810 & 0.690 & 0.745 & 26.4 & 17.0 & 0.726 & 0.661 & 0.596 & 0.527 \\
        blip2-flan-t5-xxl & 0.777 & 0.721 & 0.748 & 26.0 & 14.6 & 0.734 & 0.662 & 0.597 & 0.521 \\
        llama-adapter-v2 & 0.118 & 0.099 & 0.108 & 94.5 & 78.5 & 0.075 & 0.042 & 0.031 & 0.024 \\
        \bottomrule
    \end{tabular}
    \caption{OCR results on Memotion and Hateful Memes. We report macro precision ($\mathrm{P}_{\mathrm{macro}}$), macro recall ($\mathrm{R}_{\mathrm{macro}}$), macro F1 ($\mathrm{F1}_{\mathrm{macro}}$), word error rate (WER), character error rate (CER), and BLEU-1/2/3/4~\cite{papineni2002bleu}.}
    \label{tab:ocr}
\end{table*}

\noindent \textbf{Text Generation.}
We use n-gram-based metrics including BLEU~\cite{papineni2002bleu} ROUGE~\cite{lin2004rouge},  METEOR~\cite{lavie2004significance}, and n-gram similarity~\cite{kondrak2005n}. These metrics evaluate the MLLMs by measuring the lexical overlap between the generated text and the reference text. 
Meanwhile, we use two established similarity metrics based on pretrained language models, including BERTScore~\cite{zhang2019bertscore} and sentence embedding similarity~\cite{reimers2019sentence}, to measure the high-level semantic overlap between two answers. 
Specifically, BERTScore leverages contextualized word embeddings to capture a token's usage in a sentence and encode sequence information. 
Sentence embedding similarity $\mathrm{sim}_{\mathrm{sent}}$ is defined as the cosine similarity between the sentence embeddings of two answers: 
\begin{equation}
\mathrm{sim}_{\mathrm{sent}}\left(\mathbf{s}_i, \mathbf{s}_j\right) = \frac{\mathbf{s}_i \cdot \mathbf{s}_j}{\|\mathbf{s}_i\| \|\mathbf{s}_j\|}, ~\label{eq:SentenceSimilarity}
\end{equation}
where $\mathbf{s}_i$ is the embedding of the $i$-th response. Additionally, we calculate the length of response, defined as the number of words in a model-generate response.

\subsection{Details on Models}
\label{app:model_detail}
Table~\ref{tab:models} contains the names and number of parameters of the language encoder and vision encoder for each of the models used in our study. Table~\ref{tab:overallAccuracy} contains the accuracy scores of every classification task in our benchmark, examined across all of the zero-shot MLLMs.
\begin{table*}[!ht]
    \centering
    \begin{tabular}{lll}
    \toprule
        Model & Language Encoder & Vision Encoder \\ 
        \midrule
        llava-v1.5-7b & LLaMA-2-7B-Chat & CLIP ViT-L/14 (0.43B) \\ 
        llava-v1.5-13b & LLaMA-2-13B-Chat & CLIP ViT-L/14 (0.43B) \\ 
        instructblip-vicuna-7b & Vicuna-7B & EVA-ViT-G (1.3B) \\ 
        instructblip-vicuna-13b & Vicuna-13B & EVA-ViT-G (1.3B) \\ 
        instructblip-flan-t5-xxl & Flan-T5-XXL (11.3B) & EVA-ViT-G (1.3B) \\ 
        blip2-opt-2.7b & OPT-2.7B & EVA-ViT-G (1.3B) \\ 
        blip2-flan-t5-xxl & Flan-T5-XXL (11.3B) & EVA-ViT-G (1.3B) \\ 
        llama-adapter-v2 & LLaMA-7B & CLIP ViT-L/14 (0.43B) \\ 
        \bottomrule
    \end{tabular}
    \caption{Multimodal large language models (MLLMs) we evaluated in the experiment. }
    \label{tab:models}
\end{table*}

\begin{table*}[!ht]
    \centering
    \begin{tabular}{l|cccccc|c}
    \toprule
    Model & Misinfo & Hate & Humor & Sarc. & Off. & Sent. & Avg. \\ 
    \midrule
    llava-v1.5-7b & \underline{0.776} & 0.526 & 0.763 & 0.721 & 0.492 & 0.485 & \underline{0.627} \\
    llava-v1.5-13b & \textbf{0.800} & 0.580 & 0.767 & \underline{0.775} & \underline{0.591} & 0.327 & \textbf{0.640} \\
    instructblip-vicuna-7b & 0.319 & 0.534 & \underline{0.771} & 0.638 & 0.481 & \underline{0.547} & 0.549 \\
    instructblip-vicuna-13b & 0.494 & 0.550 & \textbf{0.776} & \underline{0.775} & \textbf{0.599} & 0.443 & 0.606 \\
    instructblip-flan-t5-xl & 0.765 & 0.508 & 0.226 & 0.560 & 0.393 & 0.387 & 0.473 \\
    instructblip-flan-t5-xxl & 0.486 & \underline{0.587} & 0.762 & \textbf{0.777} & 0.393 & 0.471 & 0.579 \\
    blip2-opt-2.7b & 0.268 & 0.508 & 0.543 & 0.393 & 0.418 & \textbf{0.637} & 0.461 \\
    blip2-flan-t5-xl & 0.770 & 0.500 & 0.224 & 0.597 & 0.393 & 0.373 & 0.476 \\
    blip2-flan-t5-xxl & 0.775 & \textbf{0.600} & 0.767 & 0.674 & 0.393 & 0.420 & 0.605 \\
    llama-adapter-v2 & 0.613 & 0.548 & 0.721 & 0.770 & 0.473 & 0.455 & 0.597 \\ 
    \midrule
    random & 0.500 & 0.500 & 0.510 & 0.502 & 0.499 & 0.326 & 0.473 \\
    \bottomrule
    \end{tabular}
    \caption{Accuracy of all models on the classification tasks. Best and 2nd best performances among the MLLMs are highlighted in \textbf{bold} and \underline{underline}, respectively. 
    }
    \label{tab:overallAccuracy}
\end{table*}

\begin{figure*}[!t]
    \centering
    \includegraphics[width=0.98\linewidth]{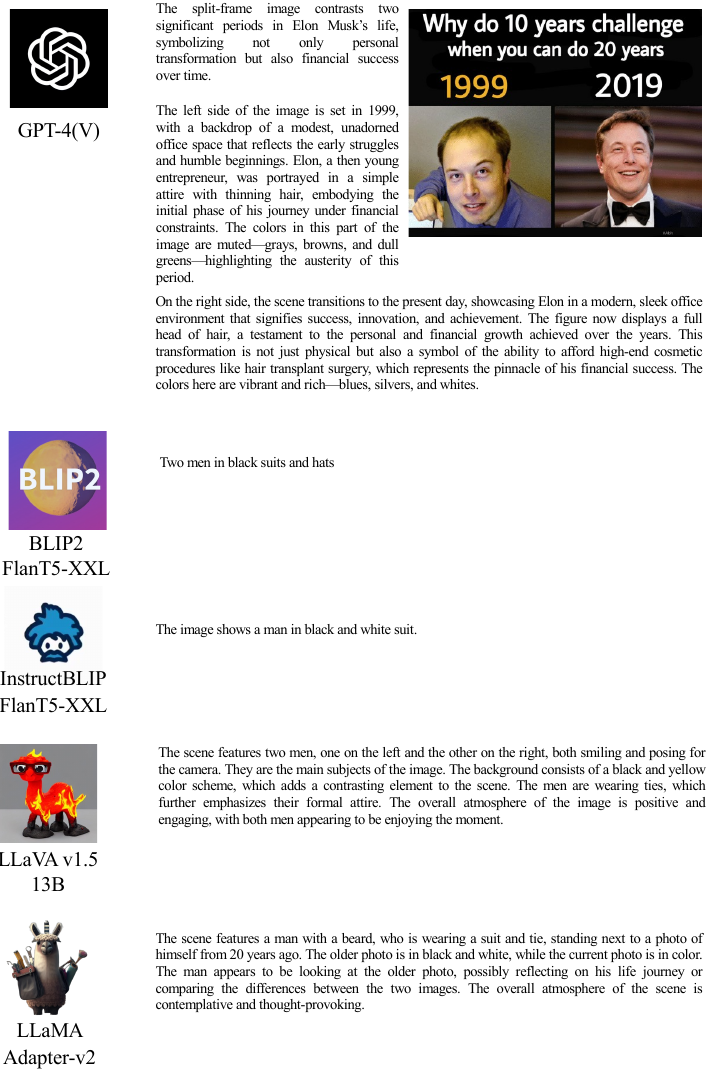}
    \caption{Example generation by GPT-4(V) and the four strongest MLLMs under each model architecture. Answers from InstructBLIP and BLIP2 are succinct, whereas those from LLaVA and LLaMA-Adapter-v2 are more comprehensive. }
    \label{fig:exampleGeneration}
    \vspace{-3mm}
\end{figure*}




\end{document}